%% file: neurips_2026.tex
\documentclass{article}
\PassOptionsToPackage{final}{graphicx}
\PassOptionsToPackage{final}{graphics}

\PassOptionsToPackage{numbers,square}{natbib}
\usepackage[main,final]{neurips_2026}
\usepackage{latexml}

\usepackage[utf8]{inputenc}
\usepackage[T1]{fontenc}    
\usepackage[table,xcdraw]{xcolor}
\usepackage{colortbl}
\usepackage{caption} 
\usepackage{url}            
\usepackage{booktabs}       
\usepackage{amsfonts}       
\usepackage{nicefrac}       
\usepackage{microtype}      
\usepackage{amsmath,amssymb}
\usepackage{graphicx}
\usepackage{wrapfig}
\usepackage{mathtools}
\usepackage{algorithm}
\usepackage{algorithmic}
\usepackage{amsthm}
\usepackage{hhline} 
\usepackage{tabularx} 
\usepackage{arydshln}
\usepackage{multirow}
\usepackage{mathrsfs} 
\usepackage{hyperref}
\hypersetup{
    colorlinks=true,
    linkcolor=blue,
    citecolor=blue,
    urlcolor=blue
}
\usepackage{bm}
\usepackage{array}
\usepackage{makecell}
\usepackage{subfigure}
\usepackage{tikz}
\usepackage{enumitem}
\usepackage{fontawesome5}
\input{preamble}

\title{CLaW: Codec-Guided Adaptive Latent Watermarking  \\ for Traceable Diffusion Image Generation}

\iflatexml
\author{
  Huangchao Song \\
  {University of Science and Technology of China} \\
  \href{mailto:songhuangchao@mail.ustc.edu.cn}{\texttt{songhuangchao@mail.ustc.edu.cn}}
  \And
  Lingyun Yu\footnotemark \\
  {Tianjin University}
  \And
  Cong Li \\
  {University of Science and Technology of China}
  \And
  Peiqi Jiang \\
  {University of Science and Technology of China}
  \And
  Mingqi Fang \\
  {University of Science and Technology of China}
  \And
  Hongtao Xie\footnotemark \\
  {University of Science and Technology of China}
}
\else
\author{
  Huangchao Song\textsuperscript{1} \enspace
  Lingyun Yu\textsuperscript{2}\thanks{Corresponding author.} \quad
  Cong Li\textsuperscript{1} \enspace
  Peiqi Jiang\textsuperscript{1} \enspace
  Mingqi Fang\textsuperscript{1} \enspace
  Hongtao Xie\textsuperscript{1}\footnotemark[1] \\
  \textsuperscript{1}University of Science and Technology of China \\
  \textsuperscript{2}Tianjin University \\
  \texttt{songhuangchao@mail.ustc.edu.cn}
}
\fi

\begin{document}
\iflatexml
  \renewcommand{\thefootnote}{✉}
  \maketitle
  \footnotetext{Corresponding author.}
\else
  \renewcommand{\fnsymbol}[1]{\text{\scalebox{0.9}{\faEnvelope}}}
  \maketitle
\fi
\setcounter{footnote}{0}

\input{sections/0_abs}
\input{sections/1_intro}
\input{sections/2_related}
\input{sections/3_method}
\input{sections/4_exp}
\input{sections/5_conclusion}

\clearpage
\begin{ack}
This work is supported by the Fundamental and Interdisciplinary Disciplines Breakthrough Plan of the Ministry of Education of China (JYB2025XDXM103), and the National Natural Science Foundation of China (62425114, 62121002, U23B2028, 62472395).
\end{ack}

\bibliographystyle{unsrt}
\bibliography{reference}

\input{sections/6_supplementary_material}

\clearpage
\input{checklist}

\end{document}

%% file: preamble.tex
\usepackage{bm}
\usepackage{amsfonts}       %
\usepackage{amsmath,amssymb}
\usepackage{mathtools}
\usepackage{algorithm}
\usepackage{algorithmic}
\usepackage{amsthm}
\usepackage{mathrsfs} %
\usepackage{amsmath} %
\usepackage{booktabs} %
\graphicspath{{figures/}{figures/pipeline/}{figures/appendix/}}
\DeclareGraphicsExtensions{.pdf,.png,.jpg,.jpeg}
\AtBeginDocument{\setkeys{Gin}{draft=false}}

\input{math_commands.tex}

\usepackage{cuted}

%% file: math_commands.tex
\usepackage{amsmath,amsfonts,bm}

\def\1{\bm{1}}

\def\vb{{\bm{b}}}

\def\vx{{\bm{x}}}

\def\vz{{\bm{z}}}

\DeclareMathAlphabet{\mathsfit}{\encodingdefault}{\sfdefault}{m}{sl}
\SetMathAlphabet{\mathsfit}{bold}{\encodingdefault}{\sfdefault}{bx}{n}

\newcommand{\softmax}{\mathrm{softmax}}

%% file: sections/0_abs.tex
\begin{abstract}

With the rapid advancement of text-to-image diffusion models, increasingly realistic AI-generated content has raised serious concerns about misuse and copyright infringement. 
Digital watermarking offers a promising solution by enabling user-level traceability, yet existing methods remain difficult to deploy at scale, hindered by costly model retraining or detection, poor effectiveness–fidelity trade-offs, and limited robustness to image transformations. 
To address these challenges, we propose \textbf{CLaW} (\textbf{C}odec-guided \textbf{La}tent \textbf{W}atermarking), a robust and efficient watermarking framework for traceable diffusion image generation with frozen diffusion backbones and inversion-free detection.
Specifically, CLaW first uses a pretrained watermark codec to map each watermark message to a latent residual, enabling low-cost sampling-time injection. 
Then, the residual is injected within a late denoising window, where image semantics are better preserved and the watermark signal is less perturbed by subsequent denoising updates, yielding a better balance between watermark effectiveness and image fidelity. 
Furthermore, we introduce a decoder-guided adaptive injection mechanism that uses decoding confidence as feedback to dynamically calibrate watermark strength, reinforcing weak watermark signals and improving robustness under image transformations. 
Extensive experiments show that CLaW preserves high visual fidelity and achieves state-of-the-art robustness under image transformations, enabling reliable user-level traceability.
\end{abstract}

%% file: sections/1_intro.tex
\section{Introduction} \label{sec:introduction}
Diffusion models~\cite{ho2020ddpm,song2021scorebasedsde} have become a dominant paradigm for high-quality image generation and are now widely deployed through commercial APIs and creative tools, including Stable Diffusion~\cite{rombach2022ldm,podell2023sdxl,esser2024rft}, DALL-E~\cite{ramesh2022dalle2,betker2023dalle3}, Imagen~\cite{saharia2022imagen}, and recent frontier image generators~\cite{google2026nanobanana2,bytedance2026seedream5lite,openai2026gptimage2}.
While these systems make AI-generated content increasingly realistic, they also raise serious concerns about synthetic-media misuse~\cite{brundage2018maliciousai,zellers2019neuralfake}, copyright infringement~\cite{somepalli2023diffusionart,somepalli2023copying}, and training-data memorization~\cite{wen2024memorization}.
As generated images are shared, edited, and redistributed beyond the service boundary, watermarking has become increasingly important for image provenance. 
It embeds imperceptible yet recoverable signals into generated content, enabling source tracing and ownership protection.

\begin{figure*}[t]
\centering
\includegraphics[width=\textwidth,height=0.30\textheight,keepaspectratio]{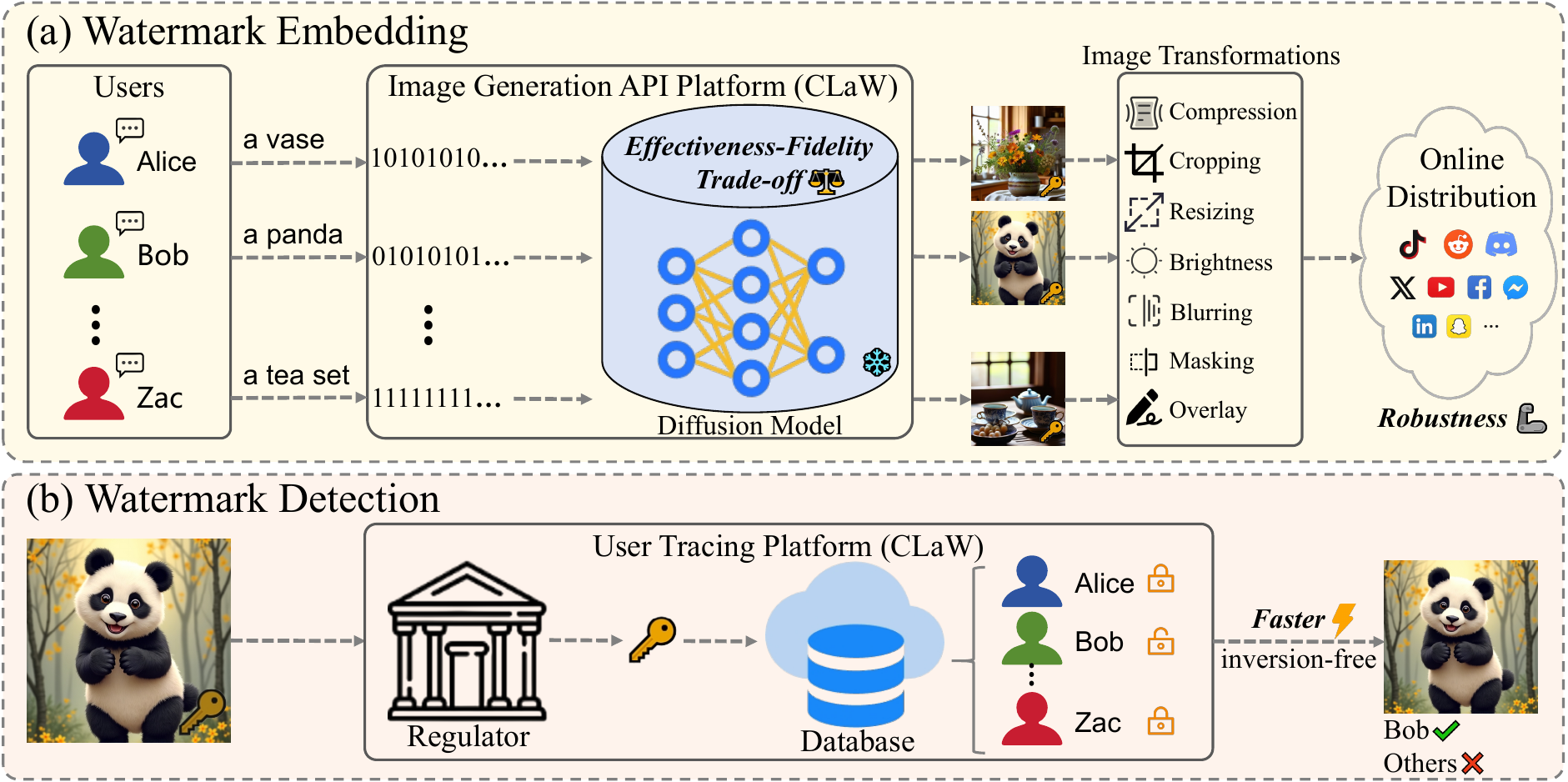}
\caption{\textbf{User-level traceability in image-generation services.}
(a) During generation, the platform assigns each user a unique watermark message and embeds it into the generated image, which may undergo common image transformations. 
(b) During detection, the watermark message is decoded from a suspicious image and matched against a registered-user database for attribution.}
\label{fig:user-trace}
\vspace{-1.0em}
\end{figure*}

Although watermarking can verify whether an image is AI-generated and identify its source model or service, achieving finer-grained user-level traceability remains a critical objective. 
We therefore focus on \emph{user-level traceability}, where each registered user is assigned a unique watermark message that can be reliably recovered from redistributed images for attribution.
Figure~\ref{fig:user-trace} illustrates this process. 
During generation, the platform embeds a user-specific watermark into the generated image. 
During detection, the watermark is decoded from a redistributed image for attribution.

However, existing watermarking methods often struggle to satisfy these requirements simultaneously.
(1) \textbf{High training or detection cost.}
Diffusion-native watermarking methods integrate watermarks directly into the generation process. Stable Signature~\cite{fernandez2023stablesignature_iccv} fine-tunes the latent decoder so that generated images carry a recoverable watermark.
Subsequent works~\cite{xiong2023fsw,min2025wadiff_eccv,ci2025wmadapter} train watermark-specific modules to better integrate watermarking with diffusion generators.
However, these approaches often require retraining the diffusion model.
Besides, sampling-time methods~\cite{wen2023treering,yang2024gaussianshading} avoid retraining but often require diffusion inversion for detection, leading to high test-time cost. 
(2) \textbf{Poor watermark effectiveness--fidelity trade-off.}
Methods such as RingID~\cite{ci2024ringid} and WIND~\cite{arabi2025wind} improve watermark effectiveness by modifying the initial noise.
Yet stronger signals alter the denoising trajectory and degrade image quality, while weaker signals fade during denoising, leading to an effectiveness--fidelity imbalance.
(3) \textbf{Limited robustness under image transformations.}
Learned post-generation watermarking methods~\cite{zhu2018hidden,tancik2020stegastamp,zhang2019rivagan} train an encoder to hide messages in the generated image and a decoder to recover them from the image.
Since the watermark message is added after generation rather than embedded during denoising, these methods are often less robust to image transformations.

To address these challenges, we propose CLaW, a robust and efficient framework for traceable diffusion image generation.
First, we pretrain a watermark codec that converts each user-specific message into a latent residual and decodes the message from the generated image. The codec is then reused with a frozen diffusion backbone, enabling low-cost sampling-time injection and inversion-free detection.
Second, to improve the watermark effectiveness--fidelity trade-off, the codec-derived residual is injected only within a late denoising window. This preserves image semantics while reducing attenuation of the watermark signal along the denoising trajectory.
Third, a decoder-guided adaptive injection mechanism is designed to selectively reinforce weak watermark signals during generation. 
Instead of using a fixed strength for all generations, CLaW uses intermediate decoding confidence as feedback to increase watermark strength when needed, thereby improving robustness.

Extensive experiments on SD3.5-medium and FLUX.1-dev demonstrate the effectiveness of CLaW.
CLaW achieves the highest average F1 under image transformations among the methods evaluated on each backbone, reaching \textbf{0.9922} and \textbf{0.9968}, respectively, while preserving the best PSNR among watermarked methods.
In summary, our main contributions are:
\begin{itemize}[leftmargin=2em, itemsep=2pt, topsep=2pt, parsep=0pt, partopsep=0pt]
\item A novel codec-guided latent watermarking framework, called CLaW, is proposed for user-level traceability in diffusion image generation, enabling low-cost sampling-time injection with frozen diffusion backbones and inversion-free detection.
\item A late-window latent injection strategy is designed to improve the effectiveness--fidelity trade-off, preserving image semantics while retaining the watermark signal through denoising.
\item A decoder-guided adaptive injection strategy is proposed for confidence-aware watermark strength calibration. It uses intermediate decoding confidence as feedback to reinforce weak watermark signals, which improves robustness under image transformations.
\end{itemize}

%% file: sections/2_related.tex
\section{Related Work} \label{sec:related_works}

\paragraph{Post-hoc image watermarking.}
Image watermarking methods can be roughly divided into two groups:
transform-based methods and learned image-domain methods.
The first group embeds signals into spatial or frequency coefficients and usually targets robustness against compression, filtering, and geometric distortions.
Early spread-spectrum and synchronization-based methods established the basic robustness framework~\cite{cox1996spread,oruanaidh1997rst}, and later studies refined practical designs and evaluation protocols for image watermarking under common signal processing operations~\cite{kutter1999benchmark,langelaar2000overview,chang2005svd,alhaj2007dwtdct,cox2007book}.
The second group learns the embedding and decoding process end-to-end. 
Generating Steganographic Images~\cite{hayes2017steganographic} and HiDDeN~\cite{zhu2018hidden} showed that neural encoders and decoders can optimize multi-bit hiding jointly with robustness objectives;
StegaStamp~\cite{tancik2020stegastamp} targets physical distortions caused by screen-shooting and print-shooting scenarios;
RivaGAN~\cite{zhang2019rivagan} adapts attention-based decoding to improve invisible video watermarking;
Certified Neural Network Watermarks~\cite{bansal2022certifiedwatermarks} and supervised GAN watermarking~\cite{fei2022supervisedganwm} further study certifiable robustness and IP protection under stronger attacks. 
However, these methods are applied after generation, so the watermark signal remains external to the diffusion process and is vulnerable to image transformations such as compression, editing, and regeneration.
This motivates CLaW to embed user-specific signals during denoising, using late-window injection and decoder-guided adaptation to improve robustness.

\paragraph{In-generation diffusion watermarking.}
Diffusion watermarking methods embed the watermark inside the diffusion pipeline so that it becomes part of the generation process. 
One branch injects the watermark into the sampling trajectory.
Tree-Ring~\cite{wen2023treering} embeds a structured pattern into the initial noise and detects it through DDIM inversion;
RingID~\cite{ci2024ringid} extends this idea to multi-key identification; 
Gaussian Shading~\cite{yang2024gaussianshading} preserves the sampling distribution while injecting watermarks; 
PRC~\cite{gunn2025prc}, WIND~\cite{arabi2025wind}, SFW~\cite{lee2025sfw}, and GaussMarker~\cite{li2025gaussmarker} further improve robustness or undetectability through refined noise-space coding and detection designs;
SERUM~\cite{kociszewski2026serum} embeds watermark patterns in the initial noise and trains a lightweight detector for inversion-free detection;
Shallow Diffuse~\cite{li2025shallowdiffuse} instead exploits low-dimensional subspaces in the generation process to decouple watermarking from content synthesis.
These methods improve robustness by coupling the watermark with the sampling process, but often rely on costly diffusion inversion for detection.
Another branch modifies decoder or conditioning components so that watermarked images can be decoded directly in the image domain. 
Stable Signature~\cite{fernandez2023stablesignature_iccv} fine-tunes the latent decoder to emit a recoverable signature;
FSW~\cite{xiong2023fsw} adds a message matrix to the decoder path to support flexible message updates;
WaDiff~\cite{min2025wadiff_eccv} treats the watermark as an additional conditioning signal for user-specific identification;
WMAdapter~\cite{ci2025wmadapter} introduces a plug-in adapter for content-adaptive watermark control.
Despite enabling direct image-domain detection, these methods often require retraining or watermark-specific modules, which increases training cost and may shift the original generation distribution.
Therefore, CLaW injects codec-derived watermark signals during sampling without modifying the diffusion backbone, reducing training cost, and uses late-window injection to balance watermark effectiveness and image fidelity.

%% file: sections/3_method.tex
\section{Method} \label{sec:method}
\begin{figure*}[t]
\centering
\includegraphics[
    width=\linewidth,
    trim=0 0.02cm 0 0,
    clip
]{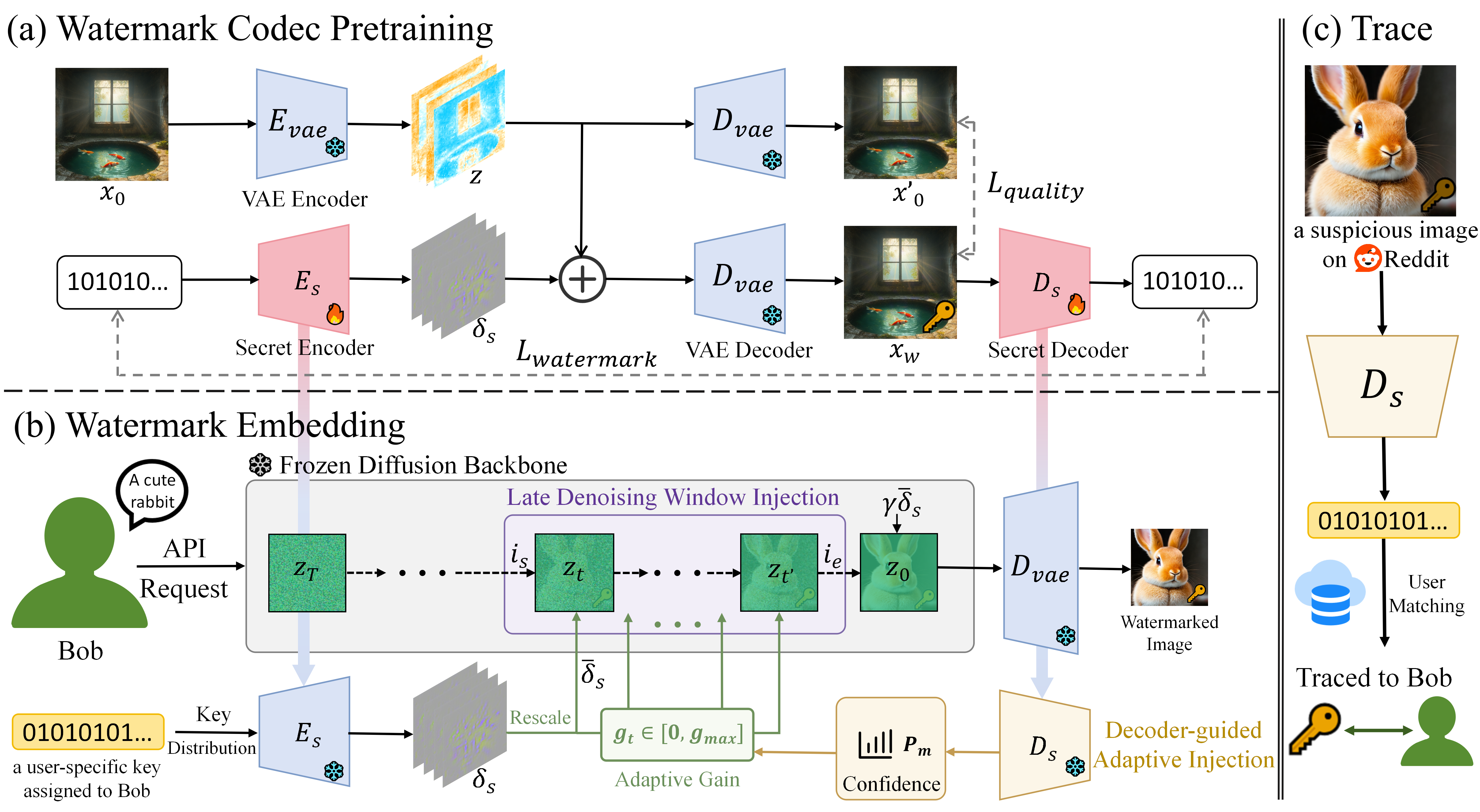}
\captionsetup{width=\linewidth}
\caption{Overview of CLaW. 
(a) Pretraining a secret encoder $E_s$ and a secret decoder $D_s$ within the latent space of a frozen VAE: $E_s$ converts a user-specific payload into a latent residual, while $D_s$ recovers the payload from the decoded image.
(b) During generation, CLaW keeps the diffusion backbone frozen and injects the residual in a late denoising window. The frozen decoder previews intermediate noisy latents and provides feedback to update the adaptive gain, which dynamically calibrates the watermark injection strength.
(c) During detection, $D_s$ directly extracts the payload from a suspicious image and matches it against the registered user database for attribution.}
\label{fig:method}
\end{figure*}

This section describes CLaW, a codec-guided latent watermarking method for traceable diffusion generation.
Each registered user is assigned a unique binary trace key. 
We refer to the embedded and recoverable binary representation as the watermark payload.
As shown in Figure~\ref{fig:method}, CLaW first pretrains a secret codec  that maps a user-specific payload to a latent residual and decodes the payload from an image (Section~\ref{sec:codec_pretraining}). 
During generation, CLaW reuses the pretrained encoder to inject the watermark residual in a late denoising window (Section~\ref{sec:late_window_injection}).
The paired decoder further guides adaptive injection by previewing payload recoverability from intermediate noisy latents and increasing the injection strength when the watermark is not yet confidently recoverable (Section~\ref{sec:adaptive_injection}).
After generation, CLaW detects the embedded payload from a suspicious image and matches it against a registered user database for attribution (Section~\ref{sec:detection_attribution}).
\subsection{Watermark Codec Pretraining}
\label{sec:codec_pretraining}
CLaW first pretrains a watermark codec $(E_s,D_s)$ following the latent-watermarking
paradigm of AquaLoRA~\cite{feng2024aqualora}.
The codec contains a secret encoder and decoder. 
Given a clean image $\vx_0$ and a payload $\vb=(b_1,\ldots,b_K)\in\{0,1\}^{K}$, the encoder maps the payload to a latent residual $\delta_s=E_s(\vb)$.
We encode the image as $\vz=E_{\mathrm{vae}}(\vx_0)$ and define its VAE reconstruction as $\vx_0^{\prime}=D_{\mathrm{vae}}(\vz)$.
The watermark residual $\delta_s$ is then embedded into the clean latent to obtain a watermarked latent $\vz+\delta_s$, which is decoded into the watermarked image: $\vx_w=D_{\mathrm{vae}}(\vz+\delta_s)$. 
The codec is trained to recover the payload after an augmentation layer \(\mathcal{A}\), 
where \(\sigma(\cdot)\) denotes the logistic sigmoid function that maps decoder logits to bit-wise probabilities.
The watermark loss is defined as:
\begin{equation}
\mathcal{L}_{\mathrm{WM}}
=
-\frac{1}{2K}\sum_{k=1}^{K}
\left[
\log \sigma(s_{k,b_k})
+
\log\left(1-\sigma(s_{k,1-b_k})\right)
\right],
\;\;
s_k = D_s(\mathcal{A}(\vx_w))_k .
\end{equation}

To make the watermark residual visually imperceptible, we pretrain the codec
with the watermark loss and two visual regularizers: LPIPS~\cite{zhang2018lpips}
and the Peak Regional Variation Loss (PRVL) proposed in
AquaLoRA~\cite{feng2024aqualora}. We optimize the following objective:
\begin{equation}
    \mathcal{L}
    =
    \mathcal{L}_{\mathrm{WM}}
    +
    \lambda\,\mathcal{L}_{\mathrm{LPIPS}}(\vx_0^{\prime}, \vx_w)
    +
    \mu\,\mathcal{L}_{\mathrm{PRVL}}(\vx_0^{\prime}, \vx_w).
    \label{eq:codec_loss}
\end{equation}
Once pretrained, $E_s$ and $D_s$ remain fixed throughout CLaW.
The frozen encoder converts each user payload into a latent watermark residual that is embedded along the generation trajectory,
while the decoder provides feedback for adaptive injection and recovers the payload during detection.
\subsection{Late-Denoising Window Watermark Injection}
\label{sec:late_window_injection}
During generation, CLaW requires no additional training and keeps both the diffusion backbone and the pretrained codec frozen.
The watermark payload is embedded in latent space by injecting codec residuals into selected noisy latents.
Given a prompt $c$ and a payload $\vb$, the secret encoder maps $\vb$ to a latent residual $\delta_s=E_s(\vb)$. 
To align with the scheduler latent scale, we rescale the VAE-space residual as \(\bar{\delta}_s = s_{\mathrm{vae}} \, \delta_s\), where $s_{\mathrm{vae}}$ is the VAE scaling factor and $\bar{\delta}_s$ denotes the watermark residual in scheduler units.
The VAE shift is not applied to the residual, as the residual is additive.
When necessary, we resize $\delta_s$ to match the latent resolution using bilinear interpolation before scaling.

We inject the watermark during a late denoising window with flexible injection schedules to balance robustness and image quality.
Let $N$ denote the number of inference steps, and let $i = 0, \ldots, N-1$ index the denoising steps. We define the injection window parameterized by two fractions $\rho_s$ and $\rho_e$ with \(i_s = \lfloor \rho_s N \rfloor\), \(i_e = \lceil \rho_e N \rceil\), and \(\mathcal{W} = \{ i_s, \ldots, i_e - 1 \}\). Watermark injection is applied within the late denoising window \(\mathcal{W}\). At each step \(i\), we define the injected residual as:
\begin{equation}
\delta_i(\vb)
=
\begin{cases}
\alpha\, w_i\, \bar{\delta}_s, & i \in \mathcal{W}, \\
0, & \text{otherwise}.
\end{cases}
\label{eq:late_window_delta}
\end{equation}
where \(\alpha\) denotes watermark strength, \(w_i\) is the normalized schedule weight at step \(i\) within the injection window \(\mathcal{W}\), and \(\bar{\delta}_s\) is the scheduler-space watermark residual.

We inject the watermark by adding the residual defined in Eq.~\eqref{eq:late_window_delta} to the latent at each step, i.e., \(\tilde{\vz}^{(i)} = \vz^{(i)} + \delta_i(\vb)\). 
The sampler then updates the watermarked latent as:
\begin{equation}
\vz^{(i+1)} = S_{\theta}\!\left(\tilde{\vz}^{(i)}, \tau_i, c \right),\;\;\;\; i = 0, \ldots, N-1.
\end{equation}
where \(S_{\theta}\) denotes the denoising update at timestep \(\tau_i\), and \(c\) is the text condition.

The final watermarked image is decoded as \(\vx_w = D_{\mathrm{vae}}(\vz^{(N)})\). 
Late injection enables robust payload embedding while preserving generation quality, as watermarks injected at early denoising steps can degrade semantic fidelity and be attenuated by the long denoising trajectory.
After the final denoising step, we optionally apply a small terminal anchor \(\gamma \bar{\delta}_s\) to the latent, providing a final reinforcement of the watermark signal.
We further introduce flexible injection schedules (e.g., flat, linear, cosine) to control the temporal distribution of watermark strength within the injection window.
Implementation details of the terminal anchor and injection schedules are provided in Appendix~\ref{app:late_window_schedule}.

\subsection{Decoder-Guided Adaptive Watermark Injection}
\label{sec:adaptive_injection}
A fixed schedule applies the same base strength $\alpha w_i$ to every prompt and every intermediate state.
For user-level traceability, this leads to inefficient signal allocation, assigning unnecessary strength to easy samples while providing insufficient signal for hard cases.
Some trajectories become reliably decodable after only a few late-stage injections, whereas others require stronger residual signals to remain recoverable under post-generation perturbations.
To address this variability, we augment the fixed rule in Eq.~\eqref{eq:late_window_delta} with a decoder-guided feedback mechanism.

We implement the feedback loop illustrated in Figure~\ref{fig:method}(b).
At selected steps, CLaW decodes the current latent using the VAE decoder and uses the pretrained secret decoder to estimate payload decodability. 
This preview acts purely as a generation-time control signal, with small additional cost and without changing the standard generation pipeline.
Formally, at a preview step $i \in \mathcal{W}$, the current latent \(\vz^{(i)}\) is decoded into a preview image $\vx_{\mathrm{pre}}^{(i)} = D_{\mathrm{vae}}(\vz^{(i)})$, where scheduler-specific latent rescaling is omitted for notational simplicity. 
Given the secret decoder $D_s$, we define the preview payload confidence
for the target payload $\vb$ as:
\begin{equation}
P_m(i)
=
\frac{1}{K}\sum_{k=1}^{K}
\softmax\!\left(D_s(\vx_{\mathrm{pre}}^{(i)})_k\right)_{b_k}.
\label{eq:preview_confidence}
\end{equation}

Here, $D_s(\vx)_k$ denotes the two-class logits for the $k$-th payload bit, and $P_m(i)$ averages the probability assigned to the target payload
$\vb=(b_1,\ldots,b_K)$. 
This soft confidence provides the feedback signal for adaptive gain control along the denoising trajectory.

CLaW adapts the injection gain by comparing the current preview confidence $P_m(i)$ with a target confidence $\tau$. 
Let $\lambda$ be the gain coefficient, and let $[a]_+=\max(a,0)$ denote the positive part operator.
To control the overhead of preview decoding, we introduce a preview stride $s_{\mathrm{pre}}$, so that $P_m(i)$ and the gain are updated only at selected steps within the injection window. 
With $i_s$ denoting the first step in $\mathcal{W}$, these preview-update steps are defined as $\mathcal{P}_{\mathrm{pre}} = \{\, i\in\mathcal{W}\mid (i-i_s)\bmod s_{\mathrm{pre}}=0 \,\}$. 
Initialized at the default value $g=1$, the gain is updated as:
\begin{equation}
    g_i
    =
    \begin{cases}
    \mathrm{clip}_{[g_{\min},g_{\max}]}
    \!\left(1+\lambda[\tau-P_m(i)]_+\right),
    & i\in\mathcal{P}_{\mathrm{pre}},\\
    g_{i^-}, & i\in\mathcal{W}\setminus\mathcal{P}_{\mathrm{pre}},
    \end{cases}
    \label{eq:adaptive_gain}
\end{equation}
Here $g_{i^-}$ denotes the last available gain before step $i$, which is reused at non-preview steps in $\mathcal{W}$.
At preview-update steps, the gain increases only when $P_m(i)<\tau$, with the increase determined by the confidence gap $\tau-P_m(i)$.
The clipping range $[g_{\min},g_{\max}]$ bounds the gain to prevent excessive amplification of the watermark residual. 
The scheduled residual is then adaptively scaled as:
\begin{equation}
    \delta_i^{\mathrm{ada}}(\vb)
    =
    \begin{cases}
    \alpha\, w_i\, g_i\, \bar{\delta}_s, & i \in \mathcal{W}, \\
    0, & \text{otherwise}.
    \end{cases}
    \label{eq:adaptive_injection_rule}
\end{equation}
Here, $w_i$ follows the same normalized late-window schedule as in Eq.~\eqref{eq:late_window_delta}. 
At each injection step, the watermarked latent is formed as
$\tilde{\vz}^{(i)}=\vz^{(i)}+\delta_i^{\mathrm{ada}}(\vb)$ and passed to the standard denoising update. 
Final payload recovery is performed after generation by a single forward pass of the frozen secret decoder on the generated image.
\subsection{Watermark Detection and User Attribution}
\label{sec:detection_attribution}

As illustrated in Figure~\ref{fig:method}(c), CLaW performs user tracing
directly from a suspicious image $\bar{\vx}$. 
The frozen secret decoder $D_s$ predicts bit-wise logits from which the payload estimate $\hat{\vb}$ is obtained.
For a candidate payload $\vb$, we define the match count as $M(\vb,\hat{\vb})=\sum_{k=1}^{K}\mathbb{I}[\hat{b}_k=b_k]$ and the bit accuracy as $\operatorname{BitAcc}(\vb,\hat{\vb})=M(\vb,\hat{\vb})/K$.
Given a registered user set $\mathcal{U}$ with payloads
$\{\vb^{(u)}\}_{u\in\mathcal{U}}$, user attribution is performed by maximizing
the payload match count:
\begin{equation}
    \hat{u}
    =
    \arg\max_{u\in\mathcal{U}} M(\vb^{(u)},\hat{\vb}).
    \label{eq:user_attribution}
\end{equation}
The attribution is accepted only if the best match satisfies
$M(\vb^{(\hat{u})},\hat{\vb})\ge\kappa$, where $\kappa$ is chosen by the binomial false-positive criterion in Appendix Eq.~\eqref{eq:bg_fpr} to ensure a low false attribution rate. 
Unlike binary watermark detection, user tracing requires the decoded payload to match a registered user-specific payload with sufficient accuracy.

%% file: sections/4_exp.tex
\section{Experiments}
\label{sec:experiments}

\newcommand{\notsup}{\multicolumn{1}{c}{--}}
In this section, we present a comprehensive set of experiments to evaluate CLaW as a practical user-level traceability mechanism for diffusion-generated images.
We first benchmark its watermarking performance and image quality on SD3.5-medium~\cite{sd35medium} and FLUX.1-dev~\cite{flux1dev} (Section~\ref{sec:exp_accuracy_quality}), 
and then evaluate its robustness to common image transformations and generation-configuration changes (Section~\ref{sec:exp_robustness}),
as well as composite transformations and screen-shooting distortions (Section~\ref{sec:exp_extended_robustness}).
We further analyze user-level attribution in Section~\ref{sec:exp_traceability} and module-level ablations in Section~\ref{sec:exp_discussion_ablation}. 
Additional evaluation results and qualitative examples are provided in Appendix~\ref{app:additional_results}.

\paragraph{Codec and generation setup.}
CLaW uses a $48$-bit watermark codec pretrained with the objective in
Section~\ref{sec:codec_pretraining}. 
In the main experiments, it injects the watermark in the last $10\%$ of the sampling trajectory with a linear schedule, decoder-guided gain adaptation, and a lightweight terminal anchor. 
We evaluate all methods on SD3.5-medium and FLUX.1-dev using shared $1{,}000$-image test sets and matched seeds. 
Additional codec, generation, and evaluation details are provided in Appendix~\ref{app:exp_details}.

\paragraph{Baselines.}
We build our comparison protocol on the open-source \emph{MarkDiffusion} framework~\cite{pan2025markdiffusion}, which provides a unified interface for generative watermarking evaluation.
We compare CLaW with diverse representative watermarking baselines: RivaGAN~\cite{zhang2019rivagan}, Gaussian Shading~\cite{yang2024gaussianshading}, RingID~\cite{ci2024ringid}, PRC~\cite{gunn2025prc}, WIND~\cite{arabi2025wind}, GaussMarker~\cite{li2025gaussmarker}, and SFW~\cite{lee2025sfw}.
We also include SERUM~\cite{kociszewski2026serum} in the SD3.5-medium robustness comparison reported in Table~\ref{tab:claw_attack_breakdown}.
A detailed conceptual comparison between CLaW and related watermarking methods is provided in Appendix~\ref{app:related_method_comparison}.

\paragraph{Evaluation metrics.}
We evaluate watermarking performance using F1, accuracy, and BitAcc.
F1 and accuracy are computed with watermarked images as positives and clean images as negatives, 
while BitAcc measures correctly recovered payload bits.
Robustness is assessed under image transformations. 
Image quality is measured with CLIP~\cite{hessel2021clipscore}, FID~\cite{heusel2017fid}, PSNR~\cite{huynhthu2008psnr}, and SSIM~\cite{wang2004ssim} following MarkDiffusion~\cite{pan2025markdiffusion}.
For user-level traceability, trace accuracy is computed by matching decoded payloads to registered payloads under the binomial threshold in Appendix Eq.~\eqref{eq:bg_fpr}.

\paragraph{Robustness evaluation.}
We evaluate robustness from two complementary aspects: robustness to image transformations and robustness to generation-configuration changes.
For robustness to image transformations, we follow the MarkDiffusion robustness protocol, including JPEG compression, crop-and-resize and Adaptive Noise,
and extend the evaluation to composite and repeated transformations, screenshots, and screen-shooting distortions.
The standard attack configurations are provided in Table~\ref{tab:attack_parameters}.
For robustness to generation-configuration changes, we evaluate CLaW across denoising steps, CFG scale, and resolution.
Model-specific settings are provided in Appendix~\ref{app:codec_generation_setup}; 
full image-attack and regeneration-attack results are reported in Appendices~\ref{app:image_attack_results} and~\ref{app:regeneration_results}, respectively.

\subsection{Benchmarking Watermark Accuracy and Image Quality}
\label{sec:exp_accuracy_quality}

\definecolor{methodgreen}{RGB}{232, 245, 238}
\definecolor{defaultgray}{RGB}{204, 204, 204}

Table~\ref{tab:combined_effectiveness} evaluates watermark accuracy, image fidelity, and runtime for traceability. 
Since clean-image F1 is largely saturated across methods, the key distinction comes from attacked F1, fidelity, and runtime, which measure robustness after redistribution and practical verification cost.

CLaW achieves the best overall trade-off between watermark reliability and visual fidelity. 
As shown in Table~\ref{tab:combined_effectiveness}, it attains perfect clean F1 on both diffusion backbones and yields the highest attacked F1 under common image transformations. 
Notably, CLaW also achieves the highest PSNR among all watermarked methods on both backbones, indicating that its robustness gains do not come at the cost of visible image degradation. 
This demonstrates a more favorable balance between robustness and image quality than prior baselines 
and indicates that the robustness gain is not obtained by introducing stronger visible distortions. 
In contrast, several diffusion-watermarking baselines achieve high clean F1 but substantially degrade PSNR and FID; RivaGAN better preserves image quality, but is less robust under attacks and slower at detection.

Runtime is also important in practice, covering both the generation overhead of watermark embedding and the cost of subsequent detection. 
CLaW adds only a small generation overhead and uses a single frozen-decoder forward pass for detection, making it substantially faster than baseline detectors. 
Overall, these results show that CLaW improves robustness and fidelity while retaining practical generation and detection costs.

\begin{table*}[!t]
\centering
\footnotesize
\caption{Main comparison on SD3.5-medium and FLUX.1-dev, evaluated with 1,000 generated images per method and backbone. 
We report watermark detection accuracy, image fidelity, and runtime. 
F1 is measured on clean generated images, while F1 (Adv.) reports the average F1 under common post-generation image transformations.
Gen. and Det. denote mean seconds per image for watermarked generation and detection, respectively.
Runtime standard deviations are reported in Appendix Table~\ref{tab:runtime_both_backbones}, and full image-quality results are reported in Appendix Tables~\ref{tab:quality_sd35_full} and~\ref{tab:quality_flux_full}.}
\label{tab:combined_effectiveness}
\begingroup
\setlength{\tabcolsep}{2.4pt}
\renewcommand{\arraystretch}{1.08}
\resizebox{\textwidth}{!}{%
\begin{tabular}{@{}lcccccccccccc@{}}
\toprule[0.8pt]
\multirow{2}{*}{Method} &
\multicolumn{6}{c}{SD3.5-medium} &
\multicolumn{6}{c}{FLUX.1-dev} \\
\cmidrule[0.4pt](lr){2-7}\cmidrule[0.4pt](lr){8-13}
& F1 $\uparrow$ & F1 (Adv.) $\uparrow$ & PSNR $\uparrow$ & FID $\downarrow$ & Gen. $\downarrow$ & Det. $\downarrow$
& F1 $\uparrow$ & F1 (Adv.) $\uparrow$ & PSNR $\uparrow$ & FID $\downarrow$ & Gen. $\downarrow$ & Det. $\downarrow$ \\
\midrule[0.4pt]
RivaGAN~\cite{zhang2019rivagan} & 0.9914 & 0.9288 & 40.53 & \textbf{35.13} & 2.56 & 0.518
& \textbf{1.0000} & 0.9577 & 40.39 & \textbf{38.30} & 5.12 & 0.530 \\
Gaussian Shading~\cite{yang2024gaussianshading} & \textbf{1.0000} & 0.8750 & 9.99 & 35.57 & 11.40 & 1.559
& \textbf{1.0000} & 0.8656 & 10.85 & 38.40 & 14.49 & 4.703 \\
RingID~\cite{ci2024ringid} & 0.9935 & 0.6213 & 8.56 & 35.38 & 2.23 & 1.589
& 0.9950 & 0.4473 & 11.09 & 38.74 & \textbf{4.57} & 4.658 \\
PRC~\cite{gunn2025prc} & 0.9712 & 0.5642 & 8.44 & 35.90 & 2.70 & 3.444
& 0.9975 & 0.5090 & 11.06 & 38.57 & 4.70 & 6.836 \\
WIND~\cite{arabi2025wind} & 0.9901 & 0.8665 & 8.31 & 36.26 & 2.34 & 2.241
& 0.9877 & 0.8658 & 10.78 & 39.31 & 4.61 & 4.846 \\
GaussMarker~\cite{li2025gaussmarker} & \textbf{1.0000} & 0.8744 & 8.50 & 36.47 & 11.27 & 2.107
& \textbf{1.0000} & 0.8737 & 11.22 & 38.78 & 13.93 & 4.702 \\
SFW~\cite{lee2025sfw} & 0.9848 & 0.6636 & 12.06 & 37.12 & \textbf{1.87} & 1.577
& 0.9950 & 0.5351 & 13.80 & 41.96 & 4.59 & 4.696 \\
\rowcolor{methodgreen}
\textbf{CLaW (ours)} & \textbf{1.0000} & \textbf{0.9922} & \textbf{41.25} & 35.70 & 1.93 & \textbf{0.0115}
& \textbf{1.0000} & \textbf{0.9968} & \textbf{41.06} & 38.37 & 4.73 & \textbf{0.0104} \\
\bottomrule[0.8pt]
\end{tabular}}
\endgroup
\end{table*}

\subsection{Benchmarking Watermark Robustness}
\label{sec:exp_robustness}

To better understand robustness beyond the average score, we further report per-attack results on SD3.5-medium in Table~\ref{tab:claw_attack_breakdown}. 
The corresponding FLUX.1-dev results are provided in Appendix Table~\ref{tab:full_attack_flux}. 
CLaW remains robust across all perturbations, with every F1 score above $0.96$, showing that its robustness is consistent across transformations rather than limited to a few specific attacks.

The per-attack results show that prior methods remain vulnerable to specific transformations. 
Gaussian Shading, GaussMarker, and WIND are brittle under crop-and-resize despite strong performance under several other corruptions, 
while RivaGAN is more balanced but still degrades under overlay. 
In contrast, CLaW maintains strong performance under both geometric and non-geometric transformations. 
This robustness aligns with its late-denoising injection and frozen codec decoder, which avoid fragile pixel-space patterns and costly inversion-based recovery.

\begin{table*}[!t]
\centering
\footnotesize
\caption{Robustness on SD3.5-medium under common image transformations.
F1 and AUC are reported for each transformation, and Avg. denotes the mean over the eight transformations.}
\label{tab:claw_attack_breakdown}
\begingroup
\scriptsize
\setlength{\tabcolsep}{3.4pt}
\renewcommand{\arraystretch}{1.08}
\resizebox{\textwidth}{!}{%
\begin{tabular}{@{}lccccccccc@{}}
\toprule[0.8pt]
\multirow{2}{*}{\textsc{Methods}} &
\multicolumn{9}{c}{\textsc{Distortions}} \\
\cmidrule[0.4pt](lr){2-10}
& JPEG & Crop+Resize & Blur & Noise & Brightness & Mask & Overlay & Adaptive Noise & Avg. \\
\midrule[0.4pt]
\multicolumn{10}{@{}l}{\textit{F1}~$\uparrow$} \\
RivaGAN~\cite{zhang2019rivagan} & 0.9189 & 0.9501 & 0.9213 & 0.9148 & 0.9496 & 0.9909 & 0.8318 & 0.9529 & 0.9288 \\
Gaussian Shading~\cite{yang2024gaussianshading} & \textbf{1.0000} & 0.0000 & \textbf{1.0000} & \textbf{1.0000} & \textbf{1.0000} & \textbf{1.0000} & \textbf{1.0000} & \textbf{1.0000} & 0.8750 \\
RingID~\cite{ci2024ringid} & 0.5380 & 0.0060 & 0.6591 & 0.4639 & 0.9712 & 0.9401 & 0.4962 & 0.8956 & 0.6213 \\
PRC~\cite{gunn2025prc} & 0.7390 & 0.0000 & 0.0198 & 0.3333 & 0.9224 & 0.9236 & 0.7063 & 0.8694 & 0.5642 \\
WIND~\cite{arabi2025wind} & 0.9901 & 0.0271 & 0.9886 & 0.9780 & 0.9877 & 0.9877 & 0.9804 & 0.9926 & 0.8665 \\
GaussMarker~\cite{li2025gaussmarker} & \textbf{1.0000} & 0.0000 & 0.9950 & \textbf{1.0000} & \textbf{1.0000} & \textbf{1.0000} & \textbf{1.0000} & \textbf{1.0000} & 0.8744 \\
SFW~\cite{lee2025sfw} & 0.6111 & 0.0000 & 0.7896 & 0.4791 & 0.9637 & 0.9610 & 0.5765 & 0.9276 & 0.6636 \\
SERUM~\cite{kociszewski2026serum} & 0.9975 & 0.9224 & 0.9738 & 0.9945 & 0.9965 & 0.9885 & 0.9849 & 0.9995 & 0.9822 \\
\rowcolor{methodgreen}
\textbf{CLaW (ours)} & 0.9995 & \textbf{0.9970} & 0.9995 & 0.9924 & 0.9616 & 0.9985 & 0.9970 & 0.9919 & \textbf{0.9922} \\
\midrule[0.4pt]
\multicolumn{10}{@{}l}{\textit{AUC}~$\uparrow$} \\
RivaGAN~\cite{zhang2019rivagan} & 0.9250 & 0.9525 & 0.9270 & 0.9215 & 0.9520 & 0.9910 & 0.8560 & 0.9550 & 0.9350 \\
Gaussian Shading~\cite{yang2024gaussianshading} & \textbf{1.0000} & 0.5000 & \textbf{1.0000} & \textbf{1.0000} & \textbf{1.0000} & \textbf{1.0000} & \textbf{1.0000} & \textbf{1.0000} & 0.9375 \\
RingID~\cite{ci2024ringid} & 0.6840 & 0.5015 & 0.7140 & 0.6510 & 0.9720 & 0.9435 & 0.6650 & 0.9055 & 0.7546 \\
PRC~\cite{gunn2025prc} & 0.7930 & 0.5000 & 0.5050 & 0.6000 & 0.9280 & 0.9290 & 0.7730 & 0.8845 & 0.7391 \\
WIND~\cite{arabi2025wind} & 0.9900 & 0.4970 & 0.9885 & 0.9775 & 0.9875 & 0.9875 & 0.9800 & 0.9925 & 0.9251 \\
GaussMarker~\cite{li2025gaussmarker} & \textbf{1.0000} & 0.5000 & 0.9950 & \textbf{1.0000} & \textbf{1.0000} & \textbf{1.0000} & \textbf{1.0000} & \textbf{1.0000} & 0.9369 \\
SFW~\cite{lee2025sfw} & 0.7200 & 0.5000 & 0.8175 & 0.6575 & 0.9650 & 0.9625 & 0.7025 & 0.9325 & 0.7822 \\
SERUM~\cite{kociszewski2026serum} & 0.9975 & 0.9255 & 0.9745 & 0.9945 & 0.9965 & 0.9885 & 0.9850 & 0.9995 & 0.9827 \\
\rowcolor{methodgreen}
\textbf{CLaW (ours)} & 0.9995 & \textbf{0.9970} & 0.9995 & 0.9925 & 0.9630 & 0.9985 & 0.9970 & 0.9920 & \textbf{0.9924} \\
\bottomrule[0.8pt]
\end{tabular}}
\endgroup
\end{table*}

We further evaluate robustness to generation-setting changes, including denoising steps, CFG scale, and output resolution. 
As shown in Table~\ref{tab:config_robustness}, CLaW remains stable on SD3.5-medium, maintaining high attacked BitAcc with only marginal CLIP variation. 
CFG scale has the most visible effect, likely because guidance strength directly alters the denoising trajectory. 
In contrast, denoising steps and resolution cause only minor fluctuations, indicating lower sensitivity to sampling budget and image size.
Overall, CLaW remains effective across different sampling budgets, guidance strengths, and output resolutions without retraining the watermark codec.

\begin{table*}[!t]
\centering
\footnotesize
\caption{Robustness to variations in SD3.5-medium and FLUX.1-dev generation settings. We change one setting at a time from the default configuration. Default test settings are colored by green cells.}
\label{tab:config_robustness}
\begingroup
\scriptsize
\setlength{\tabcolsep}{5.4pt}
\renewcommand{\arraystretch}{1.08}
\resizebox{\textwidth}{!}{
\begin{tabular}{@{}llcccccccc@{}}
\toprule[0.8pt]
\multirow{2}{*}{Factor} & \multirow{2}{*}{Setting} &
\multicolumn{4}{c}{SD3.5-medium} &
\multicolumn{4}{c}{FLUX.1-dev} \\
\cmidrule[0.4pt](lr){3-6}\cmidrule[0.4pt](lr){7-10}
& & BitAcc $\uparrow$ & BitAcc (Adv.) $\uparrow$ & CLIP $\uparrow$ & PSNR $\uparrow$ & BitAcc $\uparrow$ & BitAcc (Adv.) $\uparrow$ & CLIP $\uparrow$ & PSNR $\uparrow$ \\
\midrule[0.4pt]
\multirow{4}{*}{Steps} & 15 & 0.9875 & 0.9497 & 0.8594 & 39.3099 & 0.9975 & 0.9700 & \textbf{0.8597} & 39.4380 \\
& 25 & 0.9913 & 0.9591 & 0.8606 & 39.3239 & 0.9958 & 0.9676 & 0.8568 & 40.2712 \\
& 40 & \textbf{0.9929} & 0.9601 & \textbf{0.8626} & 39.7695 & 0.9973 & 0.9707 & 0.8571 & 40.8469 \\
& \cellcolor{methodgreen}50 & 0.9910 & 0.9582 & 0.8625 & 40.0539 & \textbf{0.9977} & \textbf{0.9708} & 0.8572 & \textbf{41.0504} \\
\midrule[0.4pt]
\multirow{3}{*}{CFG} & 5.0 & \textbf{0.9942} & \textbf{0.9676} & \textbf{0.8635} & \textbf{40.2391} & 0.9975 & \textbf{0.9718} & 0.8569 & 40.2328 \\
& \cellcolor{methodgreen}7.5 & 0.9910 & 0.9580 & 0.8625 & 40.0539 & \textbf{0.9977} & 0.9708 & \textbf{0.8572} & \textbf{41.0504} \\
& 10.0 & 0.9904 & 0.9480 & 0.8602 & 39.6930 & 0.9960 & 0.9708 & 0.8559 & 40.0955 \\
\midrule[0.4pt]
\multirow{2}{*}{Resolution} & \cellcolor{methodgreen}512$\times$512 & \textbf{0.9910} & 0.9588 & 0.8625 & 40.0539 & \textbf{0.9977} & \textbf{0.9707} & 0.8572 & 41.0504 \\
& 1024$\times$1024 & 0.9875 & \textbf{0.9704} & \textbf{0.8653} & \textbf{40.7770} & 0.9865 & 0.9568 & \textbf{0.8586} & \textbf{42.4153} \\
\bottomrule[0.8pt]
\end{tabular}}
\endgroup
\end{table*}

\subsection{Benchmarking Robustness beyond Standard Attacks}
\label{sec:exp_extended_robustness}

To assess robustness under more complex image processing, we evaluate ten conditions on SD3.5-medium, covering composite transformations and restoration.
Composite transformations comprise three pipelines: degradation combines blur, noise, and JPEG compression; editing combines brightness adjustment, overlay, and JPEG compression; and geometric processing combines crop-and-resize with JPEG compression.
Each pipeline is evaluated with one and three repetitions.
Restoration comprises four conditions: Real-ESRGAN $4\times$ super-resolution; super-resolution followed by downsampling, evaluated with one and three repetitions; and SwinIR denoising.
For denoising, we use SwinIR pretrained for Gaussian noise with standard deviation $\sigma=15$, without adding input noise.

As shown in Table~\ref{tab:extended_robustness}, CLaW achieves the highest average F1 of $0.9066$ and performs best under degradation and geometric processing.
RivaGAN handles geometric processing relatively well but shows limited robustness to restoration, particularly denoising.
Gaussian Shading and GaussMarker outperform CLaW under editing and restoration but perform poorly under geometric processing.
These results demonstrate CLaW's robustness to diverse image attacks.

\begin{table*}[!t]
\centering
\footnotesize
\caption{Robustness on SD3.5-medium under composite transformations and restoration.
Mean F1 is reported for each group, and Avg. denotes the mean over all ten conditions.
Higher is better.}
\label{tab:extended_robustness}
\begingroup
\scriptsize
\setlength{\tabcolsep}{9.2pt}
\renewcommand{\arraystretch}{1.08}
\resizebox{\textwidth}{!}{%
\begin{tabular}{@{}lcccccc@{}}
\toprule[0.8pt]
\multirow{2}{*}{Method} &
\multicolumn{3}{c}{Composite transformations} &
\multicolumn{2}{c}{Restoration} &
\multirow{2}{*}{Avg.} \\
\cmidrule[0.4pt](lr){2-4}\cmidrule[0.4pt](lr){5-6}
& Degradation & Editing & Geometric & Super-resolution & Denoising & \\
\midrule[0.4pt]
RivaGAN~\cite{zhang2019rivagan} & 0.8877 & 0.5053 & 0.7406 & 0.3709 & 0.0392 & 0.5419 \\
Gaussian Shading~\cite{yang2024gaussianshading} & 0.9847 & \textbf{1.0000} & 0.0000 & \textbf{1.0000} & \textbf{1.0000} & 0.7969 \\
RingID~\cite{ci2024ringid} & 0.0000 & 0.0000 & 0.0000 & 0.0000 & 0.0000 & 0.0000 \\
PRC~\cite{gunn2025prc} & 0.1304 & 0.4818 & 0.0000 & 0.7950 & 0.6207 & 0.4230 \\
WIND~\cite{arabi2025wind} & 0.9783 & 0.9804 & 0.0914 & 0.9934 & 0.9901 & 0.8071 \\
GaussMarker~\cite{li2025gaussmarker} & 0.9767 & 0.9975 & 0.0000 & \textbf{1.0000} & \textbf{1.0000} & 0.7948 \\
SFW~\cite{lee2025sfw} & 0.0575 & 0.4699 & 0.0000 & 0.8420 & 0.8118 & 0.4393 \\
\rowcolor{methodgreen}
\multicolumn{1}{@{}>{\columncolor{methodgreen}[3.4pt][\tabcolsep]}l}{\textbf{CLaW (ours)}}
& \textbf{0.9873} & 0.8293 & \textbf{0.8030} & 0.9473 & 0.9848
& \multicolumn{1}{>{\columncolor{methodgreen}[\tabcolsep][3.4pt]}c@{}}{\textbf{0.9066}} \\
\bottomrule[0.8pt]
\end{tabular}}
\endgroup
\end{table*}

We further evaluate payload recovery from desktop screenshots and phone photographs of watermarked images displayed on a screen, using $100$ images per condition.
Watermark payloads are recovered directly from the captured images without manual cropping or geometric correction.
Under these conditions, CLaW achieves mean BitAcc of $99.04\%$ and $84.21\%$, respectively.
These results demonstrate CLaW's robustness to screenshot and screen-shooting distortions.

\subsection{Benchmarking User-Level Traceability}
\label{sec:exp_traceability}
User-level traceability requires recovering a user-specific payload from the generated image and matching it to a registered user, rather than only detecting watermark presence. 
We therefore evaluate only methods that decode explicit bit strings. 
Several baselines in Table~\ref{tab:combined_effectiveness} are not directly applicable, as they target binary detection, often rely on slow inversion-based verification, and do not provide an efficient payload interface for large-scale attribution.

We assign each simulated user a distinct $48$-bit payload and evaluate
attribution over registries of $10^4$, $10^5$, and $10^6$ users. 
As shown in Table~\ref{tab:combined_traceability}, CLaW remains stable as the registry size increases, maintaining high attacked trace accuracy even at the largest scale. 
In contrast, RivaGAN degrades more noticeably with larger registries, while DWT-DCT-SVD~\cite{navas2008dwt} remains accurate on clean images but becomes unreliable after attacks. 
These results indicate that scalable attribution requires stable payload recovery under image transformations, not merely clean-image detectability.

\begin{table*}[!t]
\centering
\footnotesize
\caption{User-level traceability with increasing registry sizes. Trace accuracy is computed by matching each decoded $48$-bit payload to a registry of candidate users. Higher is better.}
\label{tab:combined_traceability}
\begingroup
\scriptsize
\setlength{\tabcolsep}{3.8pt}
\renewcommand{\arraystretch}{1.08}
\begin{tabular}{@{}llcccccc@{}}
\toprule[0.8pt]
\multirow{2}{*}{Backbone} & \multirow{2}{*}{Method}
& \multicolumn{2}{c}{$10^4$ users}
& \multicolumn{2}{c}{$10^5$ users}
& \multicolumn{2}{c}{$10^6$ users} \\
\cmidrule[0.4pt](lr){3-4}\cmidrule[0.4pt](lr){5-6}\cmidrule[0.4pt](lr){7-8}
& & Trace Acc. & Trace Acc. (Adv.) & Trace Acc. & Trace Acc. (Adv.) & Trace Acc. & Trace Acc. (Adv.) \\
\midrule[0.4pt]
\multirow{3}{*}{SD3.5-medium}
& DWT-DCT-SVD~\cite{navas2008dwt} & 0.9960 & 0.5090 & 0.9960 & 0.4768 & 0.9940 & 0.4464 \\
& RivaGAN~\cite{zhang2019rivagan} & 0.9950 & 0.9231 & 0.9920 & 0.9001 & 0.9870 & 0.8570 \\
& \cellcolor{methodgreen}\textbf{CLaW (ours)}
& \cellcolor{methodgreen}\textbf{1.0000} & \cellcolor{methodgreen}\textbf{0.9861}
& \cellcolor{methodgreen}\textbf{1.0000} & \cellcolor{methodgreen}\textbf{0.9786}
& \cellcolor{methodgreen}\textbf{0.9980} & \cellcolor{methodgreen}\textbf{0.9649} \\
\midrule[0.4pt]
\multirow{3}{*}{FLUX.1-dev}
& DWT-DCT-SVD~\cite{navas2008dwt} & \textbf{1.0000} & 0.5610 & \textbf{1.0000} & 0.5251 & \textbf{1.0000} & 0.4866 \\
& RivaGAN~\cite{zhang2019rivagan} & \textbf{1.0000} & 0.9454 & 0.9990 & 0.9276 & 0.9970 & 0.8935 \\
& \cellcolor{methodgreen}\textbf{CLaW (ours)}
& \cellcolor{methodgreen}\textbf{1.0000} & \cellcolor{methodgreen}\textbf{0.9936}
& \cellcolor{methodgreen}\textbf{1.0000} & \cellcolor{methodgreen}\textbf{0.9896}
& \cellcolor{methodgreen}\textbf{1.0000} & \cellcolor{methodgreen}\textbf{0.9803} \\
\bottomrule[0.8pt]
\end{tabular}
\endgroup
\end{table*}

We further assess open-set false attribution on SD3.5-medium with $10^6$ registered users.
Thresholds are set using $6{,}000$ unwatermarked images, then held fixed for testing on another $6{,}000$ unwatermarked images and $1{,}000$ watermarked images carrying distinct $48$-bit payloads.
Here, the unwatermarked images include natural images from COCO2017~\cite{lin2014coco} and images generated by SD3.5-medium.
Across five runs, CLaW achieves a mean registry-level false-positive rate (FPR) of $0.0500\%$ and $98.5\%$ trace accuracy, with the attribution threshold set to $44$ matching bits out of $48$.
Increasing the threshold to $45$ matching bits reduces FPR to $0.0100\%$, while maintaining $97.3\%$ trace accuracy.
Overall, CLaW supports accurate attribution with low FPR at the million-user scale.

\subsection{Discussion and Ablation}
\label{sec:exp_discussion_ablation}

\begin{figure*}[!t]
\centering
\includegraphics[width=0.92\textwidth]{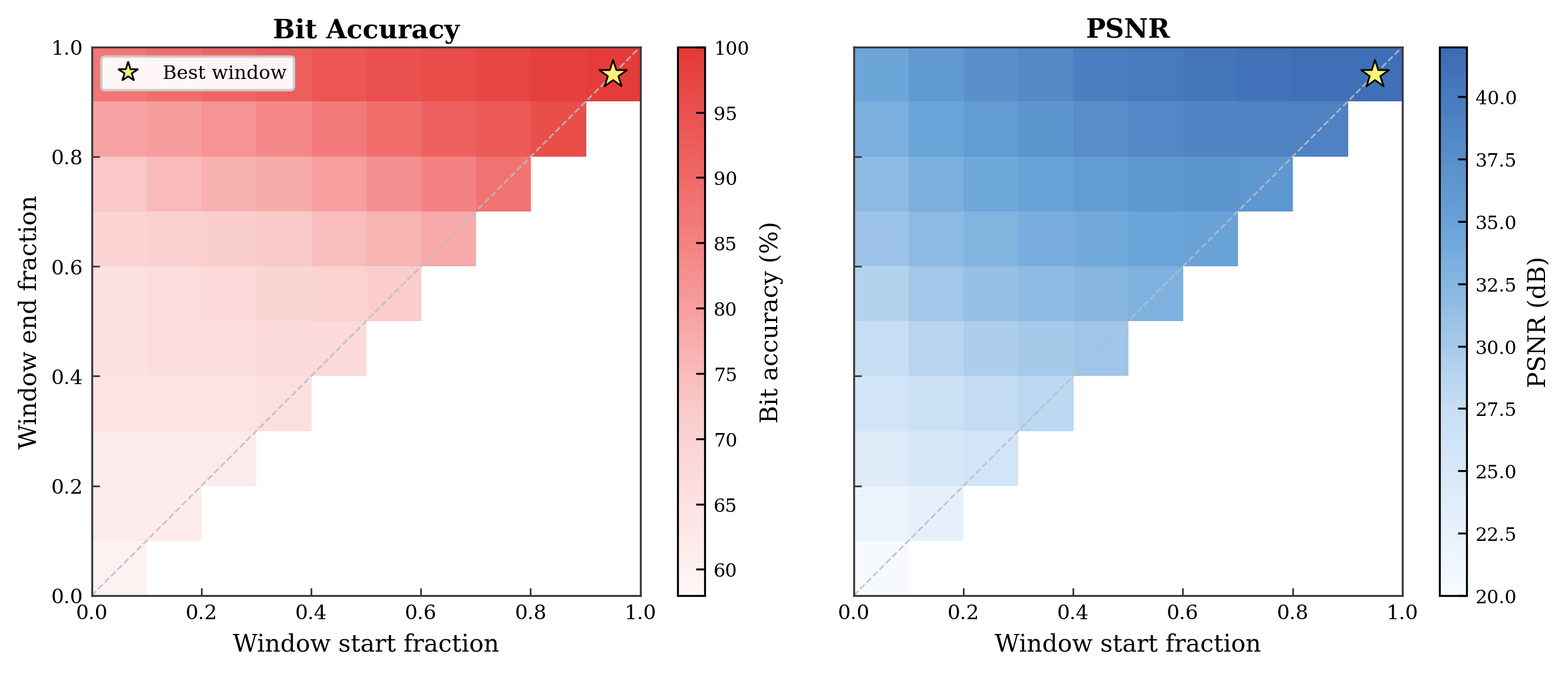}
\caption{Effect of injection timing. Cells denote denoising-window start/end pairs, with BitAcc on the left and PSNR on the right. The starred late-window setting balances recovery and fidelity.}
\label{fig:injection_timing_heatmap}
\end{figure*}

Figure~\ref{fig:injection_timing_heatmap} studies the effect of injection timing by varying the injection window along the denoising trajectory. 
High recovery is concentrated in late denoising windows, where the injected residual is less likely to be perturbed by subsequent updates. 
These windows also retain higher PSNR, showing that late-window injection balances effectiveness and fidelity. 
Additional analyses of watermark-strength and semantic consistency are provided in Appendix Figures~\ref{fig:strength_pareto} and~\ref{fig:distribution_tsne}.

Table~\ref{tab:combined_ablation} isolates the contribution of each CLaW
component. 
Removing the late denoising window sharply reduces attacked BitAcc on
both backbones, confirming that late injection is essential for payload
persistence. 
Its higher PSNR mainly reflects a weaker watermark signal, since the payload is barely embedded. Adaptive feedback and the terminal anchor provide smaller but consistent gains: adaptive feedback improves sample-specific recoverability, while the terminal anchor stabilizes the final watermark signal before decoding.

\begin{table*}[!t]
\centering
\footnotesize
\caption{Module ablation of CLaW on SD3.5-medium and FLUX.1-dev. Higher is better.}
\label{tab:combined_ablation}
\begingroup
\scriptsize
\setlength{\tabcolsep}{2.8pt}
\renewcommand{\arraystretch}{1.08}
\resizebox{\textwidth}{!}{%
\begin{tabular}{@{}lcccccccccccc@{}}
\toprule[0.8pt]
\multirow{2}{*}{Variant}
& \multicolumn{6}{c}{SD3.5-medium}
& \multicolumn{6}{c}{FLUX.1-dev} \\
\cmidrule[0.4pt](lr){2-7}\cmidrule[0.4pt](lr){8-13}
& BitAcc & BitAcc (Adv.) & AUC & AUC (Adv.) & CLIP & PSNR
& BitAcc & BitAcc (Adv.) & AUC & AUC (Adv.) & CLIP & PSNR \\
\midrule[0.4pt]
w/o Late Denoising
& 0.5595 & 0.5413 & 0.5468 & 0.5557 & \textbf{0.8702} & \textbf{57.54}
& 0.6116 & 0.5727 & 0.7095 & 0.6309 & 0.8667 & \textbf{57.8904} \\
w/o Adaptive Feedback
& 0.9825 & 0.9283 & 0.9993 & 0.9979 & \textbf{0.8702} & 42.18
& 0.9890 & 0.9384 & \textbf{1.0000} & 0.9996 & 0.8674 & 43.1592 \\
w/o Terminal Anchor
& 0.9911 & 0.9521 & \textbf{1.0000} & 0.9997 & 0.8700 & 40.62
& 0.9952 & 0.9604 & \textbf{1.0000} & \textbf{1.0000} & 0.8675 & 41.6392 \\
\rowcolor{methodgreen}
Full
& \textbf{0.9926} & \textbf{0.9595} & \textbf{1.0000} & \textbf{1.0000} & 0.8700 & 40.04
& \textbf{0.9971} & \textbf{0.9695} & \textbf{1.0000} & \textbf{1.0000} & \textbf{0.8677} & 41.0571 \\
\bottomrule[0.8pt]
\end{tabular}}
\endgroup
\end{table*}

%% file: sections/5_conclusion.tex
\section{Conclusion} \label{sec:conclusions}
In this paper, we introduce CLaW, a robust and efficient codec-guided latent watermarking framework for user-level traceability in diffusion image generation.
CLaW enables low-cost sampling-time injection with a pretrained watermark codec, without retraining the diffusion backbone.
We further design a late-window injection strategy to better balance watermark effectiveness and image fidelity, and propose a decoder-guided adaptive injection mechanism that improves robustness under image transformations.
Extensive experiments show that CLaW achieves robust user-level traceability while maintaining high visual quality across diffusion backbones.
A current limitation is that CLaW applies watermark residuals uniformly across latent channels.
Future work could explore channel-aware injection rules that place watermark signals in more robust and less perceptually sensitive channels.

%% file: sections/6_supplementary_material.tex
\clearpage
\appendix

\providecommand{\notsup}{\multicolumn{1}{c}{--}}
\providecommand{\atkcell}[3]{\makecell[c]{#1\\#2\\#3}}
\providecommand{\suppcontentssection}[2]{%
  \noindent\hypersetup{linkcolor=black}\hyperref[#1]{\textbf{\makebox[2.0em][l]{\ref*{#1}}#2}}%
  \dotfill\hyperref[#1]{\textbf{\pageref*{#1}}}\hypersetup{linkcolor=blue}\par}
\providecommand{\suppcontentssubsection}[2]{%
  \noindent\hspace{1.45em}\hypersetup{linkcolor=black}\hyperref[#1]{\makebox[3.0em][l]{\ref*{#1}}#2}%
  \dotfill\hyperref[#1]{\pageref*{#1}}\hypersetup{linkcolor=blue}\par}

\phantomsection
\pdfbookmark[1]{Supplementary Material}{supplementary-material}
\begin{center}
    {\large\bfseries CLaW: Codec-Guided Adaptive Latent Watermarking\\
    for Traceable Diffusion Image Generation}\\[0.2em]
    {\large\bfseries Appendix}
\end{center}

\vspace{1.6em}
\noindent{\bfseries Contents:}
\vspace{0.85em}

\begingroup
\small
\setlength{\parskip}{0.58em}
\suppcontentssection{app:background}{Background}
\suppcontentssubsection{app:bg_ldm}{Latent Diffusion Models}
\suppcontentssubsection{app:bg_flow_matching}{Flow-Matching Text-to-Image Backbones}
\suppcontentssubsection{app:bg_detection}{Multi-bit Watermark Detection}
\suppcontentssection{app:method_details}{Additional Algorithmic Details and Analysis}
\suppcontentssubsection{app:late_window_schedule}{Late-Denoising Window Schedule and Terminal Anchor}
\suppcontentssubsection{app:mechanism}{Mechanistic Analysis of CLaW Injection}
\suppcontentssection{app:exp_details}{Additional Experimental Details}
\suppcontentssubsection{app:codec_generation_setup}{Codec Pretraining and Generation Setup}
\suppcontentssubsection{app:metrics_attacks}{Evaluation Metrics and Image Attack Settings}
\suppcontentssection{app:additional_results}{Additional Results}
\suppcontentssubsection{app:image_attack_results}{Robustness under Image Attacks}
\suppcontentssubsection{app:regeneration_results}{Robustness under Regeneration Attack}
\suppcontentssubsection{app:efficiency_results}{Algorithmic Efficiency}
\suppcontentssubsection{app:visual_quality_results}{Visual Quality and Semantic Fidelity}
\suppcontentssubsection{app:qualitative_examples}{Qualitative Comparisons}
\suppcontentssection{app:additional_design_analysis}{Additional Algorithm Design Analysis}
\suppcontentssubsection{app:schedule_ablation}{Injection Schedule Ablation}
\suppcontentssubsection{app:strength_semantic_consistency}{Watermark Strength and Semantic Consistency}
\suppcontentssection{app:related_method_comparison}{Conceptual Comparison with Related Methods}
\endgroup

\clearpage

\input{sections/x_background}

\section{Additional Algorithmic Details and Analysis}
\label{app:method_details}

Algorithm~\ref{alg:claw} summarizes the full CLaW pipeline, including watermark injection and image-level detection.
Watermark Injection uses the frozen secret encoder to produce a payload-dependent latent residual, adds it in a late denoising window, and optionally adapts its gain using decoder-based preview confidence.
Watermark Detection decodes the payload from a suspicious image and matches it against the registered user-payload database.

\begin{algorithm}[t]
\caption{\textbf{C}odec-guided \textbf{La}tent \textbf{W}atermarking (CLaW)}
\label{alg:claw}
\small
\begin{algorithmic}[1]
\STATE \textbf{Inject watermark:}
\STATE \textbf{Input:} prompt $c$, seed $s$, payload $\vb$
\STATE \textbf{Given:} sampler $S_{\theta}$ with timesteps $\{\tau_i\}_{i=0}^{N-1}$; frozen codec $(E_s,D_s)$; VAE decoder $D_{\mathrm{vae}}$
\STATE \textbf{Hyperparameters:} $\rho_s,\rho_e,\alpha,\tau,\lambda,g_{\min},g_{\max},s_{\mathrm{pre}}$; optional terminal-anchor scale $\gamma$
\STATE \textbf{Output:} watermarked image $\vx_w$
\STATE $\vz^{(0)}\leftarrow\operatorname{Init}(s)$
\STATE $\delta_s\leftarrow E_s(\vb)$;
\STATE $\bar{\delta}_s\leftarrow s_{\mathrm{vae}}\delta_s$
\STATE $i_s\leftarrow\lfloor\rho_sN\rfloor$; $i_e\leftarrow\lceil\rho_eN\rceil$; $\mathcal{W}\leftarrow\{i_s,\ldots,i_e-1\}$
\STATE $\{w_i\}_{i\in\mathcal{W}}\leftarrow\operatorname{NormalizeSchedule}(\mathcal{W})$; $g_{i^-}\leftarrow1$
\STATE \textit{\# Late-Denoising Window Watermark Injection (Sec.~\ref{sec:late_window_injection}).}
\FOR{$i=0,\ldots,N-1$}
    \IF{$i\in\mathcal{W}$}
        \STATE \textit{\# Decoder-Guided Adaptive Watermark Injection (Sec.~\ref{sec:adaptive_injection}).}
        \IF{$(i-i_s)\bmod s_{\mathrm{pre}}=0$}
            \STATE $\vx_{\mathrm{pre}}^{(i)}\leftarrow D_{\mathrm{vae}}(\vz^{(i)})$
            \STATE $P_m(i)\leftarrow K^{-1}\sum_{k=1}^{K}\softmax(D_s(\vx_{\mathrm{pre}}^{(i)})_k)_{b_k}$
            \STATE $g_i\leftarrow\mathrm{clip}_{[g_{\min},g_{\max}]}\!\left(1+\lambda[\tau-P_m(i)]_+\right)$
        \ELSE
            \STATE $g_i\leftarrow g_{i^-}$
        \ENDIF
        \STATE $\delta_i^{\mathrm{ada}}(\vb)\leftarrow \alpha w_i g_i\,\bar{\delta}_s$
        \STATE $\tilde{\vz}^{(i)}\leftarrow\vz^{(i)}+\delta_i^{\mathrm{ada}}(\vb)$
    \ELSE
        \STATE $\tilde{\vz}^{(i)}\leftarrow\vz^{(i)}$
    \ENDIF
    \STATE $\vz^{(i+1)}\leftarrow S_{\theta}(\tilde{\vz}^{(i)},\tau_i,c)$
\ENDFOR
\STATE \textit{\# Optional terminal anchor: reinforce the watermark signal before VAE decoding.}
\IF{$\gamma>0$}
    \STATE $\vz^{(N)}\leftarrow\vz^{(N)}+\gamma\bar{\delta}_s$
\ENDIF
\STATE \textbf{return} $\vx_w =  D_{\mathrm{vae}}(\vz^{(N)})$
\STATE
\STATE \textbf{Detect watermark:}
\STATE \textbf{Input}: suspicious image $\bar{\vx}$
\STATE \textbf{Given}: frozen decoder $D_s$; payload database $\{\vb^{(u)}\}_{u\in\mathcal{U}}$; threshold $\kappa$
\STATE \textbf{Output}: decoded payload $\hat{\vb}$ and attributed user $\hat{u}$, or reject
\STATE $\hat{\vb}\leftarrow\operatorname{Decode}_{D_s}(\bar{\vx})$
\STATE $\hat{u}\leftarrow\arg\max_{u\in\mathcal{U}} M(\vb^{(u)},\hat{\vb})$
\IF{$M(\vb^{(\hat{u})},\hat{\vb})\ge\kappa$}
    \STATE \textbf{return} $\hat{\vb},\hat{u}$
\ELSE
    \STATE \textbf{return} $\hat{\vb},\mathrm{reject}$
\ENDIF
\end{algorithmic}
\end{algorithm}

\subsection{Late-Denoising Window Schedule and Terminal Anchor}
\label{app:late_window_schedule}

Let $L=|\mathcal{W}|$ and let $j(t)$ denote the local index of timestep $t\in\mathcal{W}$ in the sampling order from $\vz_T$ to $\vz_0$.
We support flat, linear, and cosine base schedules:
\begin{equation}
    \bar{w}_{j(t)} =
    \begin{cases}
    1, & \text{flat},\\
    j(t)+1, & \text{linear},\\
    1-\cos\!\left(\frac{j(t)+1}{L}\frac{\pi}{2}\right), & \text{cosine},
    \end{cases}
    \qquad
    w_t=\frac{\bar{w}_{j(t)}}{\sum_{m=0}^{L-1}\bar{w}_m}.
    \label{eq:schedule_weights}
\end{equation}
The normalized schedule keeps the overall watermark budget controlled by the global scale $\alpha$. 
After the final denoising step, CLaW optionally adds the terminal anchor $\vz_0 \leftarrow \vz_0+\gamma\Delta_{\mathcal{M}}(\vb)$, where $\gamma$ is the terminal-anchor scale. 
This update reinforces the terminal payload signal before VAE decoding.

\subsection{Mechanistic Analysis of CLaW Injection} \label{sec:mechanism}
\label{app:mechanism}

CLaW can be viewed as a controlled watermark perturbation of the latent generation trajectory. 
This subsection provides a local first-order mechanistic perspective on how this perturbation affects generation in modern DiT and flow-matching samplers. 
Let the clean conditional sampler dynamics be:
\begin{equation}
    \vz^{(i+1)}=\Phi_i(\vz^{(i)};c),\qquad i=0,\ldots,N-1,
    \label{eq:clean_sampling_dynamics}
\end{equation}
where $\Phi_i$ denotes the one-step transition map induced by the transformer prediction and the scheduler update at step $i$. 
Starting from the same initial noise, CLaW adds a payload-dependent residual to the latent before selected sampler updates:
\begin{equation}
    \tilde{\vz}^{(i)}
    =
    \hat{\vz}^{(i)}
    +
    \alpha_i\bar{\delta}_s,
    \qquad
    \hat{\vz}^{(i+1)}
    =
    \Phi_i(\tilde{\vz}^{(i)};c),
    \label{eq:perturbed_sampling_dynamics}
\end{equation}
with
\begin{equation}
    \alpha_i=
    \begin{cases}
    \alpha\,w_i\,g_i, & i\in\mathcal{W},\\
    0, & i\notin\mathcal{W}.
    \end{cases}
    \label{eq:late_injection_coeff}
\end{equation}
Here $\bar{\delta}_s$ is the scheduler-space watermark residual defined in Section~\ref{sec:late_window_injection}, 
and $\alpha_i$ is determined by the global watermark scale, the late-window schedule, and the adaptive gain.

\paragraph{VAE-to-Sampler Residual Conversion.}
The injection rule above assumes a residual expressed in the sampler latent space. 
We now describe how the codec-produced VAE-space residual is converted into this form. During codec pretraining, the secret encoder maps each payload $\vb$ to an additive residual $\delta_s=E_s(\vb)$ in the VAE latent space.
Modern latent DiT and flow-matching generators, such as SD3.5-medium and FLUX.1-dev, evolve states in a VAE-derived latent representation, 
while their samplers use model-specific latent scaling or packed layouts.
Accordingly, CLaW resizes the residual when necessary and applies the VAE scaling factor to obtain the scheduler-space residual $\bar{\delta}_s=s_{\mathrm{vae}}\delta_s$ used in Eq.~\eqref{eq:perturbed_sampling_dynamics}. 
This conversion maps the codec residual into the sampler space, allowing it to be added directly at selected denoising steps.
Its persistence through the final image depends on the remaining sampler dynamics and the chosen injection window.

\paragraph{Late-Window Watermark Signal Propagation.}
We next use a first-order expansion around the clean sampling trajectory to analyze how this sampler-space residual propagates after injection.
Let $\epsilon^{(i)}=\hat{\vz}^{(i)}-\vz^{(i)}$ denote the deviation between the watermarked and clean trajectories, which are initialized from the same noise.
Since CLaW adds the scheduler-space residual before selected sampler updates, 
the next-step deviation is driven by both the accumulated trajectory deviation and the current watermark perturbation.
A first-order expansion of Eq.~\eqref{eq:perturbed_sampling_dynamics} around the clean trajectory gives
\begin{equation}
    \epsilon^{(i+1)}
    \approx
    J_i\bigl(\epsilon^{(i)}+\alpha_i\bar{\delta}_s\bigr),
    \qquad
    J_i:=
    \frac{\partial \Phi_i(\vz^{(i)};c)}{\partial \vz^{(i)}}.
    \label{eq:linearized_perturbation_recursion}
\end{equation}
where $J_i$ denotes the local Jacobian of the one-step sampler transition evaluated at $\vz^{(i)}$, 
and $\alpha_i$ is the effective watermark strength at step $i$. 
This recursion shows how the watermark residual is propagated by the local first-order sampler dynamics.

Unrolling Eq.~\eqref{eq:linearized_perturbation_recursion} from the shared initialization $\epsilon^{(0)}=0$ gives
\begin{equation}
    \epsilon^{(N)}
    \approx
    \sum_{i=0}^{N-1}
    \left(
    \prod_{r=i}^{N-1}J_r
    \right)
    \alpha_i\bar{\delta}_s .
    \label{eq:final_perturbation_sum}
\end{equation}
This expression decomposes the terminal latent deviation into the contributions of step-wise watermark perturbations. 
Each contribution is propagated through the remaining sampler transitions after its insertion. 
Therefore, an early residual is transformed by a longer chain of Jacobians, whereas a late-window residual follows a shorter path to the terminal latent.

The terminal latent deviation is finally projected to image space by the VAE decoder. 
Using a first-order approximation around the clean terminal latent, we obtain:
\begin{equation}
    \vx_w-\vx
    \approx
    J_{\mathrm{vae}}\,\epsilon^{(N)},
    \label{eq:image_linearization}
\end{equation}
Combining this approximation with Eq.~\eqref{eq:final_perturbation_sum} gives
\begin{equation}
    \|\vx_w-\vx\|
    \le
    \|J_{\mathrm{vae}}\|
    \sum_{i=0}^{N-1}
    \left\|
    \prod_{r=i}^{N-1}J_r
    \right\|
    \alpha_i
    \|\bar{\delta}_s\|.
    \label{eq:distortion_bound}
\end{equation}
The upper bound highlights the role of injection timing: 
early residuals traverse longer chains of sampler transitions and are more likely to perturb layout, object identity, or composition, 
whereas late-window residuals follow shorter paths after the main image content has largely formed.

To relate propagation to semantic preservation, consider an idealized local decomposition into semantic and watermark-sensitive detail components. 
Under this decomposition, the local Jacobian can be written in block form as:
\begin{equation}
    J_i=
    \begin{bmatrix}
    J_i^{\mathrm{sem,sem}} & J_i^{\mathrm{sem,wm}}\\
    J_i^{\mathrm{wm,sem}} & J_i^{\mathrm{wm,wm}}
    \end{bmatrix},
    \label{eq:block_jacobian_late}
\end{equation}
where ``sem'' denotes semantic components such as layout, object identity, and composition, 
and ``wm'' denotes watermark-sensitive components relevant to payload decoding.

At late denoising steps, the main image content has largely formed, 
so semantic and watermark-sensitive components are expected to interact weakly:
\begin{equation}
    \|J_i^{\mathrm{sem,wm}}\|,\ \|J_i^{\mathrm{wm,sem}}\|
    \le
    \varepsilon_{\mathrm{late}},
    \qquad i\in\mathcal{W}.
    \label{eq:late_phase_decoupling}
\end{equation}
This approximation provides an explanatory view: 
a watermark-sensitive residual can retain payload-relevant information while causing limited semantic drift,
which helps explain why late-window injection maintains payload decodability with limited impact on the generated content.

\paragraph{Decoder-Guided Adaptive Watermark Injection.}
Since residual propagation depends on the local sampler dynamics, the effect of a fixed watermark strength varies across trajectories.
CLaW therefore uses the frozen secret decoder as a lightweight feedback module to adapt the watermark strength during generation. 
At preview steps, CLaW decodes the current latent and computes the payload confidence $P_m(i)$ in Eq.~\eqref{eq:preview_confidence}. 
For a latent $\vz$ and payload $\vb$, we write
\[
    P_m(\vz;\vb)
    =
    \frac{1}{K}\sum_{k=1}^{K}
    \softmax\!\left(D_s(D_{\mathrm{vae}}(\vz))_k\right)_{b_k}.
\]
For a small perturbation $\eta\bar{\delta}_s$, a first-order expansion gives
\begin{equation}
    P_m(\vz+\eta\bar{\delta}_s;\vb)
    \approx
    P_m(\vz;\vb)
    +
    \eta
    \left\langle
    \nabla_{\vz} P_m(\vz;\vb),
    \bar{\delta}_s
    \right\rangle ,
    \label{eq:local_confidence_view}
\end{equation}
where $\eta$ is a small scalar perturbation strength, 
$\nabla_{\vz}P_m(\vz;\vb)$ is the latent-space confidence gradient, 
and $\langle\cdot,\cdot\rangle$ denotes the inner product. 

Eq.~\eqref{eq:local_confidence_view} suggests that the residual can increase payload confidence when its induced latent update has a positive projection onto the confidence gradient.
CLaW does not compute this gradient explicitly. 
Instead, the pretrained codec provides a payload-specific residual direction, 
and the confidence gap $\tau-P_m(i)$ serves as a lightweight feedback signal. 
When the preview confidence is sufficient, the gain remains unchanged; 
otherwise, CLaW increases the gain within the clipping range to reinforce the current trajectory. 
This adaptive rule strengthens difficult trajectories while keeping the generator frozen, without requiring confidence to increase at every step.
\section{Additional Experimental Details}
\label{app:exp_details}

\subsection{Codec Pretraining and Generation Setup}
\label{app:codec_generation_setup}

\paragraph{Watermark codec pretraining.}
CLaW uses the latent watermarking codec described in Section~\ref{sec:codec_pretraining}. 
The codec is trained on MS-COCO~\cite{lin2014coco} training images resized to $512\times512$ and is reused for all CLaW generation experiments. 
Each batch samples random $48$-bit payloads. A frozen VAE maps images to latent space, the secret encoder adds a payload residual, and the secret decoder recovers the payload from the decoded watermarked image. 
We train the secret encoder and decoder with AdamW using learning rate $1\times10^{-3}$, weight decay $1\times10^{-4}$, batch size $5$, and $40$ epochs. 
The learning rate follows a StepLR schedule with step size $2$ and decay factor $0.8$. 
The warm-up stage optimizes only the bit recovery loss, while the final stage adds LPIPS and PRVL regularization with weights $5.0$ and $1.5$, respectively. 
The augmentation layer samples identity, JPEG simulation, crop-and-resize, Gaussian blur, Gaussian noise, and color jitter. After pretraining, both $E_s$ and $D_s$ are frozen.

\paragraph{Generation and injection settings.}
We evaluate CLaW on SD3.5-medium and FLUX.1-dev, covering both standard latent and packed-latent generation pipelines. 
For generation benchmarking, we sample prompts from the Stable-Diffusion-Prompts~\cite{gustavosta2022stableprompts} dataset.
Unless otherwise specified, we assign an independent $48$-bit payload to each watermarked image. 
For both SD3.5-medium and FLUX.1-dev, we generate $512\times512$ images with $50$ denoising steps, guidance scale $7.5$, fp16 precision, and a FlowMatch-Euler scheduler.
The backbones use their corresponding generation pipelines, i.e., StableDiffusion3Pipeline for SD3.5-medium and FluxPipeline for FLUX.1-dev. 
CLaW uses the same injection configuration on both backbones: 
$\alpha=1.0$, $(\rho_s,\rho_e)=(0.9,1.0)$, a linear schedule, terminal-anchor scale $\gamma=0.1$, 
target probability $\tau=0.98$, adaptive coefficient $\lambda=2.0$, gain range $[0.75,1.5]$, and preview period $s_{\mathrm{pre}}=1$.

\subsection{Evaluation Metrics and Image Attack Settings}
\label{app:metrics_attacks}

\paragraph{Evaluation metrics.}
For binary watermark detection, we report F1, accuracy, and AUC computed from detector scores over clean and watermarked images. 
For user-level traceability, BitAcc measures the fraction of correctly recovered payload bits. 
Trace accuracy is computed by matching the decoded payload to the registered user-payload database, 
and an attribution is accepted only when the nearest match exceeds the binomial threshold in Eq.~\eqref{eq:bg_fpr}.

For visual quality, we report both distribution-level and paired image-level metrics. 
FID~\cite{heusel2017fid} measures the distributional discrepancy between generated watermarked images and the MS-COCO~\cite{lin2014coco} reference images. 
CLIPScore~\cite{hessel2021clipscore} evaluates semantic consistency between the generated image and its text prompt. 
PickScore~\cite{kirstain2023pickapic} provides a preference-based image quality score, 
while NIQE~\cite{mittal2013niqe} measures no-reference naturalness. 
For paired comparisons, LPIPS~\cite{zhang2018lpips}, PSNR, and SSIM~\cite{wang2004ssim} are computed between the watermarked image and the corresponding clean image generated with the same prompt and seed.

\paragraph{Image attack settings.}
To evaluate robustness, we apply a suite of post-generation image attacks before watermark detection and payload recovery. 
These attacks cover common frequency-domain, geometric, photometric, additive, and occlusion-based corruptions. 
Table~\ref{tab:attack_parameters} summarizes the parameter settings of each image attack and the robustness aspect it is intended to stress.

\begin{table*}[t]
\centering
\footnotesize
\caption{Post-generation image attacks used for robustness evaluation. Each attack is applied before watermark detection and payload recovery.}
\label{tab:attack_parameters}
\begin{tabular}{@{}lll@{}}
\toprule[0.8pt]
Attack & Parameter & Robustness Aspect \\
\midrule[0.4pt]
JPEG Compression & quality $=50$ & compression-induced information loss \\
Crop $+$ Resize & crop ratio $=0.8$ & geometric misalignment \\
Gaussian Blur & radius $=3$ & high-frequency attenuation \\
Gaussian Noise & noise std. $\sigma=25$ & additive pixel-level corruption \\
Brightness & factor $=1.5$ & global photometric shift \\
Mask & mask ratio $=0.18$, number of masks $=8$ & partial spatial removal \\
Overlay & $16$ random strokes, stroke width $=8$ & localized occlusion \\
Adaptive Noise & intensity $=0.9$, auto-select enabled & content-aware local corruption \\
\bottomrule[0.8pt]
\end{tabular}
\end{table*}

\begin{figure*}[!t]
\centering
\includegraphics[width=0.66\textwidth]{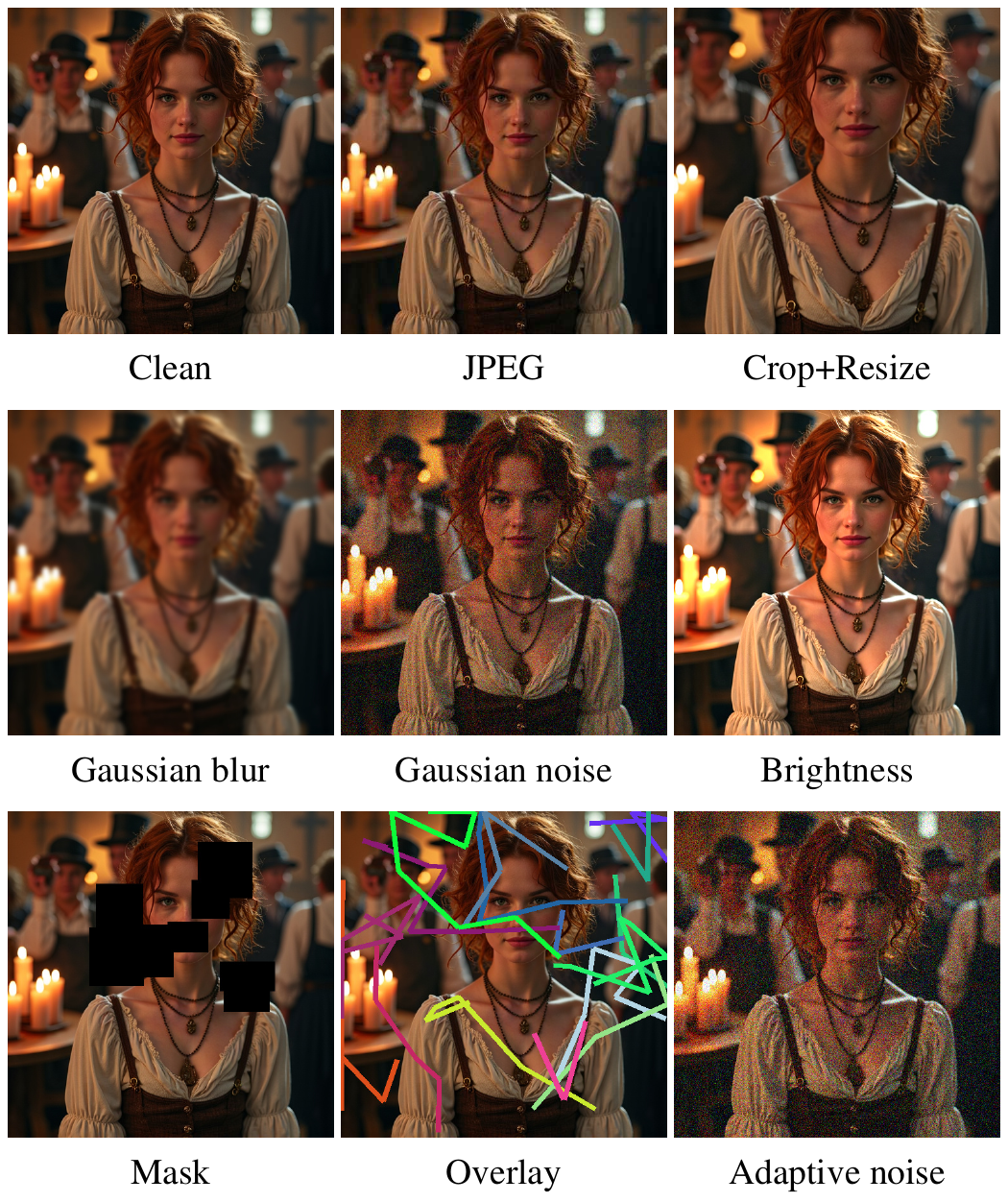}
\caption{Examples of post-generation image attacks on a FLUX.1-dev sample. 
The attacked results in the main text and supplementary tables are averaged over this perturbation suite.}
\label{fig:attack_examples_flux}
\end{figure*}

Figure~\ref{fig:attack_examples_flux} shows representative examples of the post-generation image attacks used in our robustness evaluation. 
The attack suite covers complementary robustness aspects, including compression-induced information loss, geometric misalignment, high-frequency attenuation, additive pixel-level corruption, global photometric shift, partial spatial removal, localized occlusion, and content-aware local corruption. 
Together, these attacks test watermark detectability and payload decodability under diverse image corruptions.

\section{Additional Results}
\label{app:additional_results}

\subsection{Robustness under Image Attacks}
\label{app:image_attack_results}

Tables~\ref{tab:full_attack_sd35} and~\ref{tab:full_attack_flux} report the per-attack robustness results omitted from the main text. 
Across both backbones, CLaW maintains stable detection and payload recovery across individual perturbations, 
complementing the averaged robustness results in the main tables.

\begin{table*}[t]
\centering
\scriptsize
\caption{Full attack-wise robustness on SD3.5-medium under common image transformations. Results are grouped by metric, and Avg. denotes the mean over the eight transformations.}
\label{tab:full_attack_sd35}
\begingroup
\setlength{\tabcolsep}{3.2pt}
\renewcommand{\arraystretch}{1.04}
\resizebox{\textwidth}{!}{
\begin{tabular}{@{}lccccccccc@{}}
\toprule[0.8pt]
\multirow{2}{*}{\textsc{Methods}} &
\multicolumn{9}{c}{\textsc{Distortions}} \\
\cmidrule[0.4pt](lr){2-10}
& JPEG & Crop+Resize & Blur & Noise & Brightness & Mask & Overlay & Adaptive Noise & Avg. \\
\midrule[0.4pt]
\multicolumn{10}{@{}l}{\textit{F1}~$\uparrow$} \\
RivaGAN~\cite{zhang2019rivagan} & 0.9189 & 0.9501 & 0.9213 & 0.9148 & 0.9496 & 0.9909 & 0.8318 & 0.9529 & 0.9288 \\
Gaussian Shading~\cite{yang2024gaussianshading} & \textbf{1.0000} & 0.0000 & \textbf{1.0000} & \textbf{1.0000} & \textbf{1.0000} & \textbf{1.0000} & \textbf{1.0000} & \textbf{1.0000} & 0.8750 \\
RingID~\cite{ci2024ringid} & 0.5380 & 0.0060 & 0.6591 & 0.4639 & 0.9712 & 0.9401 & 0.4962 & 0.8956 & 0.6213 \\
PRC~\cite{gunn2025prc} & 0.7390 & 0.0000 & 0.0198 & 0.3333 & 0.9224 & 0.9236 & 0.7063 & 0.8694 & 0.5642 \\
WIND~\cite{arabi2025wind} & 0.9901 & 0.0271 & 0.9886 & 0.9780 & 0.9877 & 0.9877 & 0.9804 & 0.9926 & 0.8665 \\
GaussMarker~\cite{li2025gaussmarker} & \textbf{1.0000} & 0.0000 & 0.9950 & \textbf{1.0000} & \textbf{1.0000} & \textbf{1.0000} & \textbf{1.0000} & \textbf{1.0000} & 0.8744 \\
SFW~\cite{lee2025sfw} & 0.6111 & 0.0000 & 0.7896 & 0.4791 & 0.9637 & 0.9610 & 0.5765 & 0.9276 & 0.6636 \\
\rowcolor{methodgreen}
\textbf{CLaW (ours)} & 0.9995 & \textbf{0.9970} & 0.9995 & 0.9924 & 0.9616 & 0.9985 & 0.9970 & 0.9919 & \textbf{0.9922} \\
\midrule[0.4pt]
\multicolumn{10}{@{}l}{\textit{AUC}~$\uparrow$} \\
RivaGAN~\cite{zhang2019rivagan} & 0.9250 & 0.9525 & 0.9270 & 0.9215 & 0.9520 & 0.9910 & 0.8560 & 0.9550 & 0.9350 \\
Gaussian Shading~\cite{yang2024gaussianshading} & \textbf{1.0000} & 0.5000 & \textbf{1.0000} & \textbf{1.0000} & \textbf{1.0000} & \textbf{1.0000} & \textbf{1.0000} & \textbf{1.0000} & 0.9375 \\
RingID~\cite{ci2024ringid} & 0.6840 & 0.5015 & 0.7140 & 0.6510 & 0.9720 & 0.9435 & 0.6650 & 0.9055 & 0.7546 \\
PRC~\cite{gunn2025prc} & 0.7930 & 0.5000 & 0.5050 & 0.6000 & 0.9280 & 0.9290 & 0.7730 & 0.8845 & 0.7391 \\
WIND~\cite{arabi2025wind} & 0.9900 & 0.4970 & 0.9885 & 0.9775 & 0.9875 & 0.9875 & 0.9800 & 0.9925 & 0.9251 \\
GaussMarker~\cite{li2025gaussmarker} & \textbf{1.0000} & 0.5000 & 0.9950 & \textbf{1.0000} & \textbf{1.0000} & \textbf{1.0000} & \textbf{1.0000} & \textbf{1.0000} & 0.9369 \\
SFW~\cite{lee2025sfw} & 0.7200 & 0.5000 & 0.8175 & 0.6575 & 0.9650 & 0.9625 & 0.7025 & 0.9325 & 0.7822 \\
\rowcolor{methodgreen}
\textbf{CLaW (ours)} & 0.9995 & \textbf{0.9970} & 0.9995 & 0.9925 & 0.9630 & 0.9985 & 0.9970 & 0.9920 & \textbf{0.9924} \\
\midrule[0.4pt]
\multicolumn{10}{@{}l}{\textit{BitAcc}~$\uparrow$} \\
RivaGAN~\cite{zhang2019rivagan} & 0.9383 & 0.9537 & 0.9422 & 0.9198 & 0.9565 & 0.9880 & 0.8945 & 0.9688 & 0.9452 \\
Gaussian Shading~\cite{yang2024gaussianshading} & 0.9555 & 0.4962 & 0.8608 & 0.9024 & 0.9827 & 0.9820 & 0.9632 & 0.9623 & 0.8881 \\
GaussMarker~\cite{li2025gaussmarker} & \textbf{0.9966} & 0.5118 & 0.9762 & \textbf{0.9857} & \textbf{0.9995} & \textbf{0.9995} & \textbf{0.9982} & \textbf{0.9953} & 0.9329 \\
\rowcolor{methodgreen}
\textbf{CLaW (ours)} & 0.9729 & \textbf{0.9625} & \textbf{0.9882} & 0.9303 & 0.9255 & 0.9674 & 0.9501 & 0.9754 & \textbf{0.9590} \\
\bottomrule[0.8pt]
\end{tabular}}
\endgroup
\end{table*}

\begin{table*}[t]
\centering
\scriptsize
\caption{Full attack-wise robustness on FLUX.1-dev under common image transformations. Results are grouped by metric, and Avg. denotes the mean over the eight transformations.}
\label{tab:full_attack_flux}
\begingroup
\setlength{\tabcolsep}{3.2pt}
\renewcommand{\arraystretch}{1.04}
\resizebox{\textwidth}{!}{
\begin{tabular}{@{}lccccccccc@{}}
\toprule[0.8pt]
\multirow{2}{*}{\textsc{Methods}} &
\multicolumn{9}{c}{\textsc{Distortions}} \\
\cmidrule[0.4pt](lr){2-10}
& JPEG & Crop+Resize & Blur & Noise & Brightness & Mask & Overlay & Adaptive Noise & Avg. \\
\midrule[0.4pt]
\multicolumn{10}{@{}l}{\textit{F1}~$\uparrow$} \\
RivaGAN~\cite{zhang2019rivagan} & 0.9610 & 0.9899 & 0.9770 & 0.9501 & 0.9822 & \textbf{1.0000} & 0.8270 & 0.9744 & 0.9577 \\
Gaussian Shading~\cite{yang2024gaussianshading} & 0.9975 & 0.0000 & \textbf{1.0000} & 0.9529 & \textbf{1.0000} & \textbf{1.0000} & \textbf{1.0000} & 0.9744 & 0.8656 \\
RingID~\cite{ci2024ringid} & 0.0100 & 0.0000 & 0.0000 & 0.0000 & 0.9610 & 0.9583 & 0.7421 & 0.9071 & 0.4473 \\
PRC~\cite{gunn2025prc} & 0.2063 & 0.0000 & 0.0000 & 0.0296 & 0.9717 & 0.9899 & 0.9501 & 0.9247 & 0.5090 \\
WIND~\cite{arabi2025wind} & 0.9524 & 0.0645 & \textbf{1.0000} & 0.9852 & 0.9756 & 0.9804 & 0.9828 & 0.9852 & 0.8658 \\
GaussMarker~\cite{li2025gaussmarker} & 0.9950 & 0.0000 & \textbf{1.0000} & \textbf{0.9950} & \textbf{1.0000} & \textbf{1.0000} & \textbf{1.0000} & \textbf{1.0000} & 0.8737 \\
SFW~\cite{lee2025sfw} & 0.2379 & 0.0000 & 0.1900 & 0.0000 & 0.9873 & 0.9822 & 0.9418 & 0.9418 & 0.5351 \\
\rowcolor{methodgreen}
\textbf{CLaW (ours)} & \textbf{1.0000} & \textbf{1.0000} & \textbf{1.0000} & 0.9848 & \textbf{1.0000} & \textbf{1.0000} & \textbf{1.0000} & 0.9899 & \textbf{0.9968} \\
\midrule[0.4pt]
\multicolumn{10}{@{}l}{\textit{AUC}~$\uparrow$} \\
RivaGAN~\cite{zhang2019rivagan} & 0.9625 & 0.9900 & 0.9775 & 0.9525 & 0.9825 & \textbf{1.0000} & 0.8525 & 0.9750 & 0.9616 \\
Gaussian Shading~\cite{yang2024gaussianshading} & 0.9975 & 0.5000 & \textbf{1.0000} & 0.9550 & \textbf{1.0000} & \textbf{1.0000} & \textbf{1.0000} & 0.9750 & 0.9284 \\
RingID~\cite{ci2024ringid} & 0.5025 & 0.5000 & 0.5000 & 0.5000 & 0.9625 & 0.9600 & 0.7950 & 0.9150 & 0.7044 \\
PRC~\cite{gunn2025prc} & 0.5575 & 0.5000 & 0.5000 & 0.5075 & 0.9725 & 0.9900 & 0.9525 & 0.9300 & 0.7388 \\
WIND~\cite{arabi2025wind} & 0.9500 & 0.4925 & \textbf{1.0000} & \textbf{0.9850} & 0.9750 & 0.9800 & 0.9825 & 0.9850 & 0.9188 \\
GaussMarker~\cite{li2025gaussmarker} & 0.9950 & 0.5000 & \textbf{1.0000} & 0.9950 & \textbf{1.0000} & \textbf{1.0000} & \textbf{1.0000} & \textbf{1.0000} & 0.9363 \\
SFW~\cite{lee2025sfw} & 0.5675 & 0.5000 & 0.5525 & 0.5000 & 0.9875 & 0.9825 & 0.9450 & 0.9450 & 0.7475 \\
\rowcolor{methodgreen}
\textbf{CLaW (ours)} & \textbf{1.0000} & \textbf{1.0000} & \textbf{1.0000} & \textbf{0.9850} & \textbf{1.0000} & \textbf{1.0000} & \textbf{1.0000} & 0.9900 & \textbf{0.9969} \\
\midrule[0.4pt]
\multicolumn{10}{@{}l}{\textit{BitAcc}~$\uparrow$} \\
RivaGAN~\cite{zhang2019rivagan} & 0.9581 & \textbf{0.9783} & 0.9733 & 0.9289 & 0.9831 & 0.9972 & 0.8934 & 0.9856 & 0.9622 \\
Gaussian Shading~\cite{yang2024gaussianshading} & 0.8825 & 0.4880 & 0.8508 & 0.8014 & 0.9931 & 0.9972 & 0.9874 & 0.9577 & 0.8698 \\
GaussMarker~\cite{li2025gaussmarker} & \textbf{0.9846} & 0.4897 & 0.9772 & \textbf{0.9437} & \textbf{0.9999} & \textbf{1.0000} & \textbf{0.9999} & \textbf{0.9862} & 0.9227 \\
\rowcolor{methodgreen}
\textbf{CLaW (ours)} & 0.9725 & 0.9497 & \textbf{0.9914} & 0.9248 & 0.9832 & 0.9806 & 0.9697 & 0.9781 & \textbf{0.9688} \\
\bottomrule[0.8pt]
\end{tabular}}
\endgroup
\end{table*}

\subsection{Robustness under Regeneration Attack}
\label{app:regeneration_results}

Table~\ref{tab:regen_attack} evaluates denoising-based regeneration, a stronger attack setting that partially re-synthesizes the image beyond isolated pixel-level perturbations. 
CLaW maintains stable detection and payload recovery across both
backbones, whereas several baselines degrade substantially on FLUX.1-dev despite strong performance on SD3.5-medium. 
These results highlight the importance of backbone-agnostic robustness under regeneration attacks.

\begin{table*}[t]
\centering
\scriptsize
\setlength{\tabcolsep}{4.0pt}
\caption{Robustness to denoising-based regeneration. The attack uses strength $0.15$, $20$ regeneration steps, guidance scale $4.5$, and dataset prompts. Results are grouped by backbone.}
\label{tab:regen_attack}
\resizebox{0.73\textwidth}{!}{
\begin{tabular}{@{}lcccccc@{}}
\toprule[0.8pt]
\multirow{2}{*}{Method} &
\multicolumn{3}{c}{SD3.5-medium} &
\multicolumn{3}{c}{FLUX.1-dev} \\
\cmidrule[0.4pt](lr){2-4}\cmidrule[0.4pt](lr){5-7}
& F1 $\uparrow$ & AUC $\uparrow$ & BitAcc $\uparrow$ & F1 $\uparrow$ & AUC $\uparrow$ & BitAcc $\uparrow$ \\
\midrule[0.4pt]
RivaGAN~\cite{zhang2019rivagan} & 0.7878 & 0.8250 & 0.9094 & 0.7097 & 0.7750 & 0.8828 \\
Gaussian Shading~\cite{yang2024gaussianshading} & \textbf{1.0000} & \textbf{1.0000} & 0.9426 & 0.9899 & 0.9900 & 0.8077 \\
RingID~\cite{ci2024ringid} & 0.7879 & 0.8250 & \notsup & 0.0000 & 0.5000 & \notsup \\
PRC~\cite{gunn2025prc} & 0.7450 & 0.8000 & \notsup & 0.0000 & 0.5000 & \notsup \\
WIND~\cite{arabi2025wind} & \textbf{1.0000} & \textbf{1.0000} & \notsup & 0.9615 & 0.9600 & \notsup \\
GaussMarker~\cite{li2025gaussmarker} & \textbf{1.0000} & \textbf{1.0000} & \textbf{0.9973} & \textbf{1.0000} & \textbf{1.0000} & \textbf{0.9439} \\
SFW~\cite{lee2025sfw} & 0.9744 & 0.9750 & \notsup & 0.0400 & 0.5100 & \notsup \\
\rowcolor{methodgreen}
\textbf{CLaW (ours)} & 0.9744 & 0.9750 & 0.9396 & 0.9744 & 0.9750 & 0.9177 \\
\bottomrule[0.8pt]
\end{tabular}}
\end{table*}

\subsection{Algorithmic Efficiency}
\label{app:efficiency_results}

Table~\ref{tab:runtime_both_backbones} compares the algorithmic efficiency of different watermarking methods. CLaW adds only minor generation overhead and enables millisecond-level detection with a single frozen-decoder forward pass, avoiding costly inversion.

\newcommand{\pmcell}[2]{\makebox[3.8em][r]{#1}$\,\pm\,$\makebox[3.2em][l]{#2}}
\newcommand{\notsupcell}{\makebox[3.8em][r]{}--\makebox[3.2em][l]{}}
\begin{table*}[t]
\centering
\scriptsize
\setlength{\tabcolsep}{4.0pt}
\caption{Generation and detection efficiency on an H100 80G GPU. We report mean runtime with standard deviation. The detection time is measured per image after generation. Lower is better.}
\label{tab:runtime_both_backbones}
\resizebox{0.99\textwidth}{!}{
\begin{tabular}{@{}lcccc@{}}
\toprule[0.8pt]
\multirow{2}{*}{Method} &
\multicolumn{2}{c}{SD3.5-medium} &
\multicolumn{2}{c}{FLUX.1-dev} \\
\cmidrule[0.4pt](lr){2-3}\cmidrule[0.4pt](lr){4-5}
& Gen. time (s) & Det. time (s)
& Gen. time (s) & Det. time (s) \\
\midrule[0.4pt]
RivaGAN~\cite{zhang2019rivagan} & \pmcell{2.5617}{0.3592} & \pmcell{0.5177}{0.0294} & \pmcell{5.1161}{0.1984} & \pmcell{0.5303}{0.0369} \\
Gaussian Shading~\cite{yang2024gaussianshading} & \pmcell{11.4012}{0.9618} & \pmcell{1.5594}{0.2518} & \pmcell{14.4887}{0.7680} & \pmcell{4.7032}{0.0428} \\
RingID~\cite{ci2024ringid} & \pmcell{2.2270}{0.3949} & \pmcell{1.5889}{0.2624} & \pmcell{4.5675}{0.0191} & \pmcell{4.6583}{0.0238} \\
PRC~\cite{gunn2025prc} & \pmcell{2.7029}{0.2068} & \pmcell{3.4440}{0.4512} & \pmcell{4.6993}{0.0314} & \pmcell{6.8361}{0.8122} \\
WIND~\cite{arabi2025wind} & \pmcell{2.3406}{0.3686} & \pmcell{2.2410}{0.4159} & \pmcell{4.6119}{0.0427} & \pmcell{4.8455}{0.2438} \\
GaussMarker~\cite{li2025gaussmarker} & \pmcell{11.2716}{0.8649} & \pmcell{2.1072}{0.5018} & \pmcell{13.9295}{0.6462} & \pmcell{4.7024}{0.0410} \\
SFW~\cite{lee2025sfw} & \pmcell{\textbf{1.8679}}{0.2650} & \pmcell{1.5767}{0.2165} & \pmcell{\textbf{4.5857}}{0.1466} & \pmcell{4.6961}{0.0340} \\
Clean & \pmcell{1.8341}{0.1897} & \notsupcell & \pmcell{4.6072}{0.1803} & \notsupcell \\
\rowcolor{methodgreen}
\textbf{CLaW (ours)} & \pmcell{1.9339}{0.1142} & \pmcell{\textbf{0.0115}}{0.0047} & \pmcell{4.7331}{0.0251} & \pmcell{\textbf{0.0104}}{0.0056} \\
\bottomrule[0.8pt]
\end{tabular}}
\end{table*}

\subsection{Visual Quality and Semantic Fidelity}
\label{app:visual_quality_results}

Tables~\ref{tab:quality_sd35_full} and~\ref{tab:quality_flux_full} report full image-quality results on SD3.5-medium and FLUX.1-dev. 
CLaW preserves visual and semantic fidelity across both backbones, suggesting that its robustness is not driven by stronger visible artifacts.

\begin{table*}[t]
\centering
\scriptsize
\setlength{\tabcolsep}{4.0pt}
\caption{Full image-quality results on SD3.5-medium. Lower values indicate better performance for FID and NIQE, while higher values are better for CLIP, PickScore, PSNR, and SSIM.}
\label{tab:quality_sd35_full}
\resizebox{0.87\textwidth}{!}{
\begin{tabular}{@{}lcccccc@{}}
\toprule[0.8pt]
Method & CLIP & FID (vs. COCO) & PickScore & NIQE & PSNR & SSIM \\
\midrule[0.4pt]
RivaGAN~\cite{zhang2019rivagan} & 0.8758 & \textbf{35.1276} & 22.2388 & 10.0031 & 40.5322 & \textbf{0.9820} \\
Gaussian Shading~\cite{yang2024gaussianshading} & \textbf{0.8779} & 35.5688 & 22.2567 & 10.5437 & 9.9909 & 0.1838 \\
RingID~\cite{ci2024ringid} & 0.8755 & 35.3834 & 22.3202 & 10.0119 & 8.5606 & 0.2031 \\
PRC~\cite{gunn2025prc} & 0.8755 & 35.9001 & 22.3141 & 10.1161 & 8.4404 & 0.1960 \\
WIND~\cite{arabi2025wind} & 0.8760 & 36.2584 & 22.2901 & 10.8505 & 8.3103 & 0.1901 \\
GaussMarker~\cite{li2025gaussmarker} & 0.8760 & 36.4749 & \textbf{22.3337} & 10.5729 & 8.4976 & 0.2000 \\
SFW~\cite{lee2025sfw} & 0.8731 & 37.1225 & 22.1375 & \textbf{8.1922} & 12.0641 & 0.4232 \\
Clean & 0.8748 & 35.4536 & 22.2853 & 10.2275 & \notsup & \notsup \\
\rowcolor{methodgreen}
\textbf{CLaW (ours)} & 0.8744 & 35.6994 & 22.1668 & 10.2541 & \textbf{41.2460} & 0.9427 \\
\bottomrule[0.8pt]
\end{tabular}}
\end{table*}

\begin{table*}[t]
\centering
\scriptsize
\setlength{\tabcolsep}{4.0pt}
\caption{Full image-quality results on FLUX.1-dev. Lower values indicate better performance for FID and NIQE, while higher values are better for CLIP, PickScore, PSNR, and SSIM.}
\label{tab:quality_flux_full}
\resizebox{0.87\textwidth}{!}{
\begin{tabular}{@{}lcccccc@{}}
\toprule[0.8pt]
Method & CLIP & FID (vs. COCO) & PickScore & NIQE & PSNR & SSIM \\
\midrule[0.4pt]
RivaGAN~\cite{zhang2019rivagan} & \textbf{0.8693} & \textbf{38.3027} & 22.5825 & 9.3157 & 40.3885 & 0.9817 \\
Gaussian Shading~\cite{yang2024gaussianshading} & 0.8650 & 38.3992 & 22.5498 & 9.3590 & 10.8475 & 0.3057 \\
RingID~\cite{ci2024ringid} & 0.8634 & 38.7425 & 22.5805 & 9.5687 & 11.0946 & 0.3212 \\
PRC~\cite{gunn2025prc} & 0.8629 & 38.5748 & \textbf{22.7049} & 9.5986 & 11.0556 & 0.3060 \\
WIND~\cite{arabi2025wind} & 0.8627 & 39.3105 & 22.4199 & 9.7733 & 10.7848 & 0.3106 \\
GaussMarker~\cite{li2025gaussmarker} & 0.8650 & 38.7848 & 22.6271 & 9.4042 & 11.2247 & 0.2999 \\
SFW~\cite{lee2025sfw} & 0.8521 & 41.9550 & 22.1176 & \textbf{8.6357} & 13.8034 & 0.4437 \\
Clean & 0.8667 & 38.7804 & 22.6134 & 9.3310 & \notsup & \notsup \\
\rowcolor{methodgreen}
\textbf{CLaW (ours)} & 0.8677 & 38.3650 & 22.4992 & 9.3159 & \textbf{41.0571} & \textbf{0.9948} \\
\bottomrule[0.8pt]
\end{tabular}}
\end{table*}

\subsection{Qualitative Comparisons}
\label{app:qualitative_examples}

\begin{figure*}[p]
\centering
\makebox[\textwidth][c]{%
  \hspace*{-2.2cm}%
  \includegraphics[
    width=1.2\textwidth,
    height=1.2\textheight,
    keepaspectratio
  ]{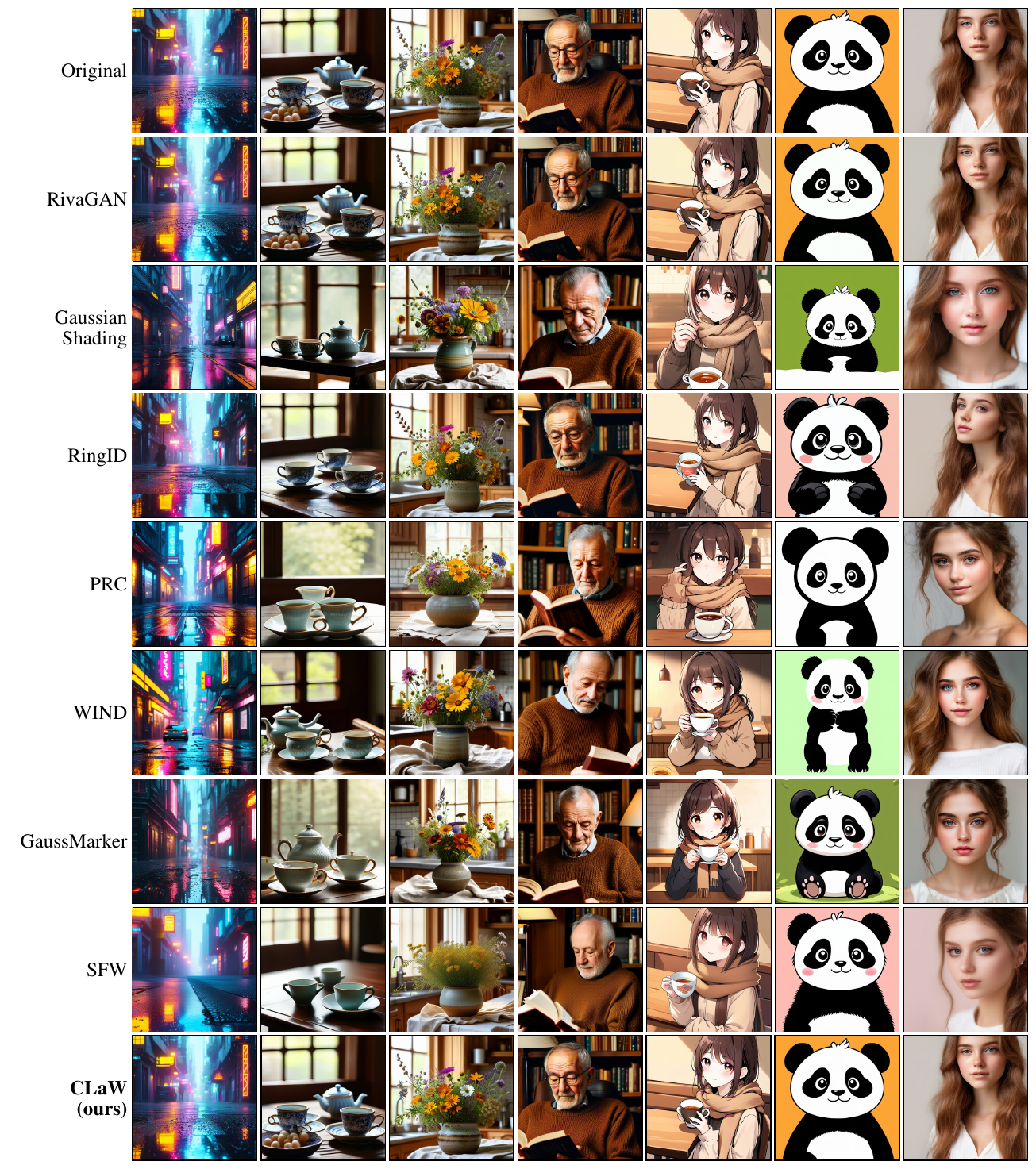}%
}
\caption{Qualitative comparison on SD3.5-medium. Rows show the clean generation, baseline watermarked outputs, and CLaW outputs. 
Columns cover representative prompt categories, including urban scenes, still-life objects, portraits, and stylized images.}
\label{fig:qualitative_sd35}
\end{figure*}

\begin{figure*}[p]
\centering
\makebox[\textwidth][c]{%
  \hspace*{-2.2cm}%
  \includegraphics[
    width=1.2\textwidth,
    height=1.2\textheight,
    keepaspectratio
  ]{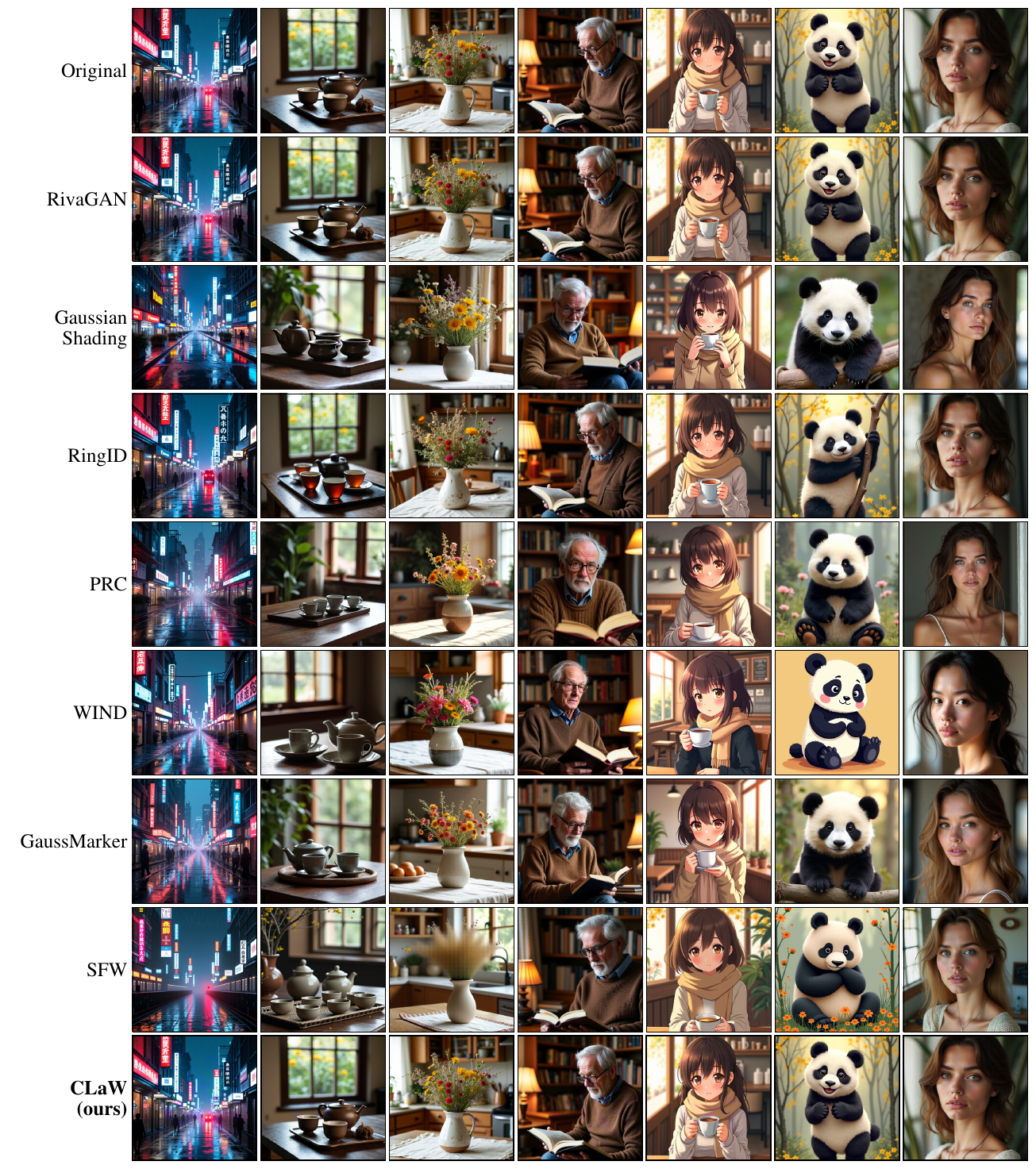}%
}
\caption{Qualitative comparison on FLUX.1-dev. Rows show the clean generation, baseline watermarked outputs, and CLaW outputs. 
Columns cover representative prompt categories, including urban scenes, still-life objects, portraits, and stylized images.}
\label{fig:qualitative_flux}
\end{figure*}

Figures~\ref{fig:qualitative_sd35} and~\ref{fig:qualitative_flux} compare clean generations, baseline watermarked outputs, and CLaW outputs on representative prompts. 
The examples complement the quantitative quality metrics: several baselines introduce visible changes in color, composition, or local texture, 
whereas CLaW more consistently preserves the appearance of the clean generation across both backbones.

\clearpage

\clearpage
\section{Additional Algorithm Design Analysis}
\label{app:additional_design_analysis}

This section provides additional analysis of CLaW's algorithmic design choices.
We first ablate the temporal schedule used for late-window injection, 
and then examine the watermark-strength operating point and a qualitative CLIP-space check for semantic consistency.

\subsection{Injection Schedule Ablation}
\label{app:schedule_ablation}

Table~\ref{tab:injection_schedule_ablation} compares flat, linear, and cosine late-window schedules across both backbones under the same injection window and adaptive controller.
Since clean F1 is saturated for all schedules, the comparison mainly reflects attacked BitAcc and image fidelity. 
Cosine gives the strongest attacked BitAcc, while linear remains close in robustness and better preserves image fidelity.
We therefore use linear by default, as it balances robustness and fidelity consistently across backbones.

\begin{table*}[t]
\centering
\footnotesize
\caption{Ablation of late-window injection schedules across backbones. 
Flat, linear, and cosine schedules are compared under the same injection window and adaptive controller.}
\label{tab:injection_schedule_ablation}
\begingroup
\setlength{\tabcolsep}{3.2pt}
\renewcommand{\arraystretch}{1.08}
\begin{tabular*}{\textwidth}{@{\extracolsep{\fill}}llccccccc@{}}
\toprule[0.8pt]
Backbone & Schedule
& F1 $\uparrow$
& F1$_{\mathrm{adv}}$ $\uparrow$
& BitAcc $\uparrow$
& BitAcc$_{\mathrm{adv}}$ $\uparrow$
& CLIP $\uparrow$
& PSNR $\uparrow$
& SSIM $\uparrow$ \\
\midrule[0.4pt]
\multirow{3}{*}{SD3.5-medium}
& cosine & 1.0000 & \textbf{0.9921} & \textbf{0.9933} & \textbf{0.9629}
& 0.8699 & \textbf{40.0530} & 0.9926 \\
& flat & 1.0000 & 0.9863 & 0.9913 & 0.9522
& \textbf{0.8700} & 39.7911 & \textbf{0.9928} \\
& linear & 1.0000 & 0.9914 & 0.9926 & 0.9598
& 0.8700 & 40.0424 & 0.9928 \\
\midrule[0.4pt]
\multirow{3}{*}{FLUX.1-dev}
& cosine & 1.0000 & \textbf{0.9972} & \textbf{0.9979} & \textbf{0.9723}
& \textbf{0.8677} & 41.0518 & 0.9947 \\
& flat & 1.0000 & 0.9959 & 0.9955 & 0.9635
& 0.8676 & 40.8491 & 0.9946 \\
& linear & 1.0000 & 0.9965 & 0.9971 & 0.9681
& 0.8677 & \textbf{41.0571} & \textbf{0.9948} \\
\bottomrule[0.8pt]
\end{tabular*}
\endgroup
\end{table*}

\subsection{Watermark Strength and Semantic Consistency}
\label{app:strength_semantic_consistency}

\begin{figure*}[t]
\centering
\subfigure[Watermark strength Pareto curve.]{
    \includegraphics[width=0.47\textwidth]{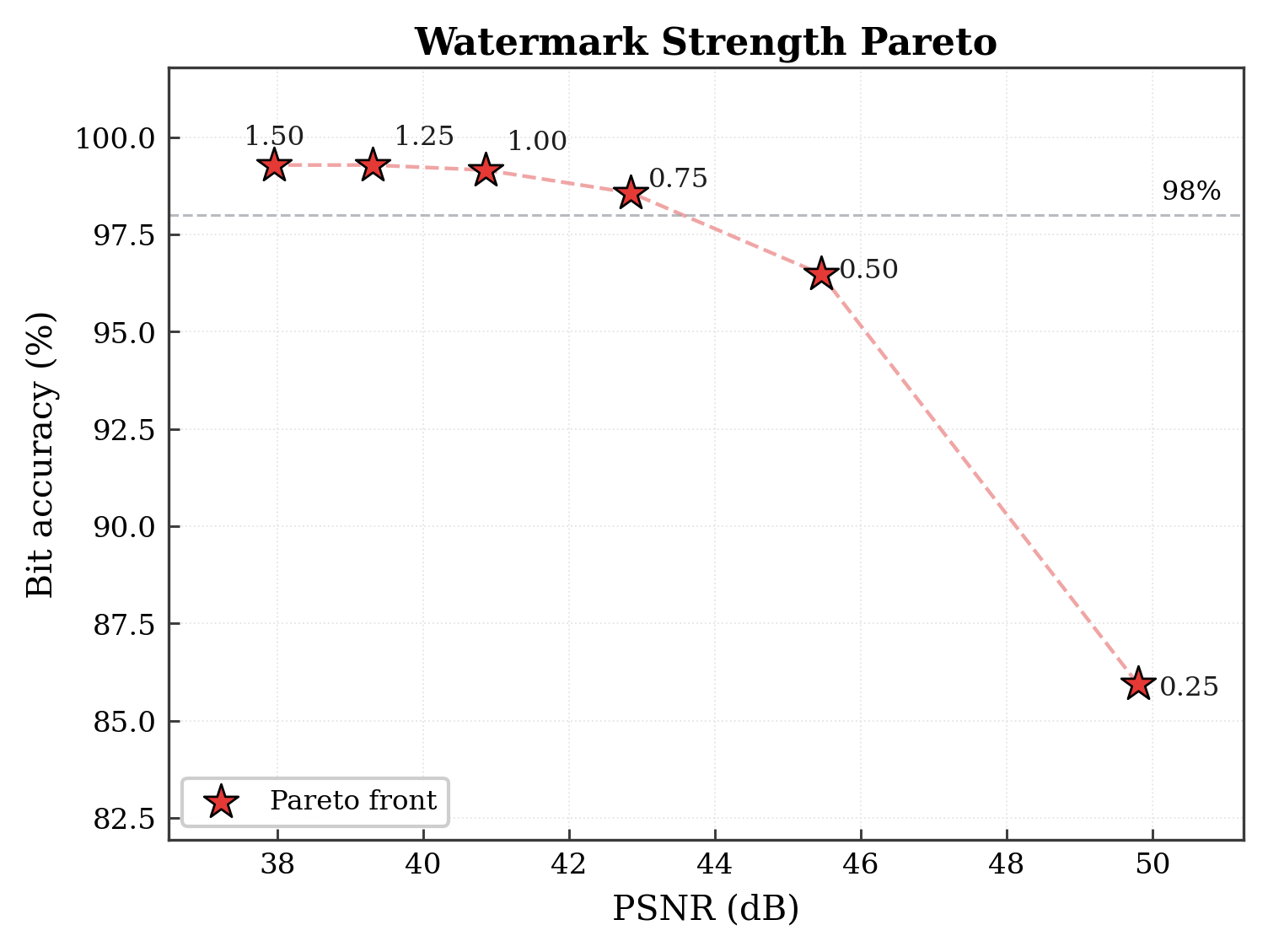}
    \label{fig:strength_pareto}}
\hfill
\subfigure[CLIP-space visualization (t-SNE).]{
    \includegraphics[width=0.47\textwidth]{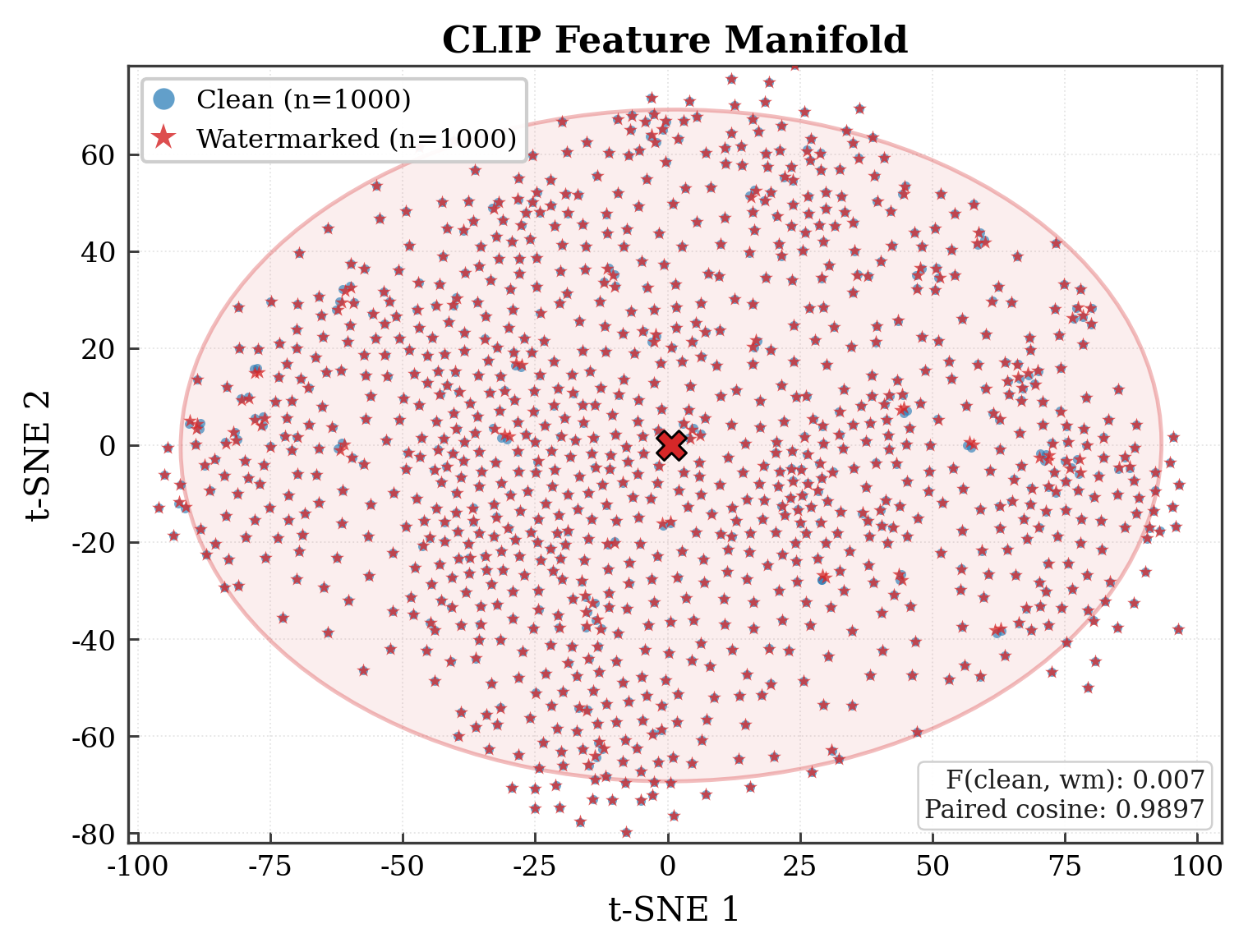}
    \label{fig:distribution_tsne}}
\caption{Additional CLaW design analysis. 
(a) PSNR--BitAcc Pareto curve under different watermark residual scales, with labels indicating tested scales.
(b) t-SNE visualization of CLIP image features for clean generated images and watermarked generated images.}
\label{fig:appendix_design_analysis}
\end{figure*}

Figure~\ref{fig:appendix_design_analysis} examines two design aspects of CLaW: 
the choice of watermark strength and a qualitative CLIP-space check for semantic consistency.
The strength sweep shows the expected trade-off: smaller residual scales better preserve PSNR but weaken payload recovery, whereas larger scales improve BitAcc with diminishing returns.
We choose the main setting near the knee of this curve, balancing recovery reliability and visual fidelity without using an unnecessarily large injection strength.
We visualize CLIP image features using t-SNE as a qualitative check. 
Clean generated images and watermarked generated images occupy nearby regions in the CLIP feature space. 
This pattern is consistent with the CLIPScore and PSNR results, suggesting that CLaW preserves semantic content and image fidelity while enabling robust and efficient user-level traceability.

\section{Conceptual Comparison with Related Methods}
\label{app:related_method_comparison}

This section provides a conceptual comparison between CLaW and related watermarking methods.
We first summarize their attribution levels, generator modifications, and watermarking mechanisms in Table~\ref{tab:conceptual_method_comparison},
and then discuss the key differences in watermark embedding and detection.

\begin{table*}[!htbp]
\centering
\footnotesize
\caption{Conceptual comparison of CLaW and related watermarking methods.}
\label{tab:conceptual_method_comparison}
\begingroup
\setlength{\tabcolsep}{4.0pt}
\renewcommand{\arraystretch}{1.15}
\renewcommand{\tabularxcolumn}[1]{m{#1}}
\begin{tabularx}{\textwidth}{lll>{\raggedright\arraybackslash}X}
\toprule[0.8pt]
Method & Attribution Level & Generator Modification & Watermarking Mechanism \\
\midrule[0.4pt]
ProMark~\cite{asnani2024promark} & Concept-level & Diffusion model training & Concept-specific watermarks learned from watermarked training images \\
\addlinespace[2pt]
Stable Signature~\cite{fernandez2023stablesignature_iccv} & Model-level & VAE decoder fine-tuning & Fixed binary payload embedding through a fine-tuned VAE decoder \\
\addlinespace[2pt]
AquaLoRA~\cite{feng2024aqualora} & Model-level & U-Net LoRA fine-tuning & Binary payload embedding through a diagonal scaling matrix in the watermark LoRA module \\
\addlinespace[2pt]
RivaGAN~\cite{zhang2019rivagan} & User-level & No diffusion/VAE update & Post-generation binary payload embedding with an attention-based encoder \\
\addlinespace[2pt]
SERUM~\cite{kociszewski2026serum} & User-level & No diffusion/VAE update & Initial-noise watermark embedding with separately trained pattern-specific detectors \\
\addlinespace[2pt]
\rowcolor{methodgreen}
\textbf{CLaW (ours)} & User-level & No diffusion/VAE update & Binary payload embedding via decoder-guided adaptive injection within a late denoising window \\
\bottomrule[0.8pt]
\end{tabularx}
\endgroup
\end{table*}

\textbf{ProMark} embeds concept-specific watermarks into training images and trains the diffusion model to retain them during generation.
This enables concept-level attribution of generated images.
CLaW instead embeds user-specific binary payloads during generation for attribution to registered users.

\textbf{Stable Signature} fine-tunes the VAE decoder to embed a fixed binary payload.
Distinct user signatures therefore require separately fine-tuned decoders.
\textbf{AquaLoRA} fine-tunes a watermark LoRA module in the U-Net and represents binary payloads through a diagonal scaling matrix, allowing payload changes without retraining.
In comparison, CLaW supports user-specific payloads through a shared pretrained watermark codec without updating the diffusion backbone or native VAE.

\textbf{RivaGAN} embeds binary payloads into the final RGB image after generation using a learned attention-based encoder.
Both RivaGAN and CLaW recover explicit binary payloads that can be matched to registered users.
CLaW instead injects a payload-dependent latent residual within a late denoising window, using intermediate decoding confidence as feedback to adapt the watermark strength.

\textbf{SERUM} embeds watermark patterns into the initial diffusion noise and trains a separate detector for each pattern.
Its multi-user extension assigns users distinct combinations of patterns and identifies them by combining the corresponding detector scores.
Both methods keep the generator fixed, train auxiliary watermarking networks, and support inversion-free detection.
In contrast to this pattern-based approach, CLaW embeds an explicit $48$-bit binary payload through a shared pretrained watermark codec and directly recovers the payload bits for matching against a registered user database.

%% file: sections/x_background.tex
\section{Background} \label{app:background}

\subsection{Latent Diffusion Models}
\label{app:bg_ldm}

Diffusion models synthesize data by learning the reverse dynamics of a gradual noising process~\cite{ho2020ddpm}. 
Latent diffusion models improve computational efficiency by transferring this denoising process from pixel space to the compact latent space of a pretrained autoencoder~\cite{rombach2022ldm}. 
An image is encoded as $\vz=E_{\mathrm{vae}}(\vx)$ and decoded as
$\vx=D_{\mathrm{vae}}(\vz)$. 
In the standard noise-prediction formulation, the forward process samples $\vz_t\sim\mathcal{N}(\sqrt{\bar{\alpha}_t}\vz_0, (1-\bar{\alpha}_t)\mathbf{I})$, 
and the denoiser is trained with:
\begin{equation}
    \mathcal{L}_{\mathrm{DM}}
    =
    \mathbb{E}_{t,\vz_0,\epsilon}
    \left\|
    \epsilon-\epsilon_{\theta}(\vz_t,t,c)
    \right\|_2^2 .
    \label{eq:bg_ddpm}
\end{equation}
At inference time, sampling starts from Gaussian noise and produces a sequence of intermediate latents, with the final latent decoded by $D_{\mathrm{vae}}$. 
CLaW is a plug-in inference-time mechanism: 
it keeps the text encoder, diffusion backbone, scheduler, and VAE frozen, 
and only injects a small payload-dependent residual into selected late denoising steps.

\subsection{Flow-Matching Text-to-Image Backbones}
\label{app:bg_flow_matching}

Our experiments use modern latent text-to-image backbones such as SD3.5 and FLUX, 
which are better described as flow-matching or rectified-flow pipelines rather than classical DDPM samplers. 
These models learn a velocity field that transports a noise latent toward a data latent. 
A common straight-line path is $\vz_t=(1-t)\epsilon+t\vz_0$, with target velocity $u_t=\vz_0-\epsilon$ and generative ODE $d\vz_t/dt=v_{\theta}(\vz_t,t,c)$. 
The corresponding flow-matching loss is
\begin{equation}
    \mathcal{L}_{\mathrm{FM}}
    =
    \mathbb{E}_{t,\vz_0,\epsilon}
    \left\|
    v_{\theta}(\vz_t,t,c)-u_t
    \right\|_2^2 .
    \label{eq:bg_flow_matching}
\end{equation}
In high-resolution text-to-image systems, the velocity field
\(v_{\theta}\) is often parameterized by a transformer over latent tokens, as in rectified-flow transformer models~\cite{esser2024rft}. 
However, CLaW does not depend on this internal parameterization. 
It only assumes access to intermediate latent states and the final VAE decoder, which are shared by both DDPM-style and flow-matching pipelines. 
Thus, CLaW operates purely at the sampler interface, injecting a
codec-generated residual into the latent trajectory without modifying transformer blocks, attention layers, or model weights.

\subsection{Multi-bit Watermark Detection}
\label{app:bg_detection}

For user-level traceability, each registered user is assigned a payload $\vb\in\{0,1\}^{K}$, 
and the detector recovers $\hat{\vb}\in\{0,1\}^{K}$ from a suspicious image. 
Let $M(\vb,\hat{\vb})=\sum_{k=1}^{K}\mathbb{I}[b_k=\hat b_k]$ be the payload match count. 
Under the null hypothesis, an unrelated image yields decoded bits that match a fixed payload independently with probability \(1/2\) per bit, 
i.e., $M\sim\mathrm{Binomial}(K,1/2)$, 
so the false-positive probability for accepting a match at threshold $m$ is:
\begin{equation}
    \mathrm{FPR}(m)
    =
    \Pr[M\ge m]
    =
    \sum_{j=m}^{K}\binom{K}{j}2^{-K}.
    \label{eq:bg_fpr}
\end{equation}
This threshold defines the attribution criterion: CLaW accepts the best registry match only when it is unlikely under the random-bit null model.

%% file: checklist.tex
\section*{NeurIPS Paper Checklist}

\begin{enumerate}

\item {\bf Claims}
    \item[] Question: Do the main claims made in the abstract and introduction accurately reflect the paper's contributions and scope?
    \item[] Answer: \answerYes{}
    \item[] Justification: 
    The paper contributes a codec-guided latent watermarking framework for traceable diffusion image generation, with the method introduced in Section~\ref{sec:method}, mechanistic analysis in  Appendix~\ref{sec:mechanism}, and experimental verification in Section~\ref{sec:experiments}.

\item {\bf Limitations}
    \item[] Question: Does the paper discuss the limitations of the work performed by the authors?
    \item[] Answer: \answerYes{}
    \item[] Justification:
    We discuss the limitation of CLaW and possible directions for future work in Section~\ref{sec:conclusions}, including its uniform latent-channel injection rule and channel-aware extensions.

\item {\bf Theory assumptions and proofs}
    \item[] Question: For each theoretical result, does the paper provide the full set of assumptions and a complete (and correct) proof?
    \item[] Answer: \answerNA{}
    \item[] Justification: 
    The paper includes mathematical descriptions of the method, but does not state any theorems or provide formal proofs.

    \item {\bf Experimental result reproducibility}
    \item[] Question: Does the paper fully disclose all the information needed to reproduce the main experimental results of the paper to the extent that it affects the main claims and/or conclusions of the paper (regardless of whether the code and data are provided or not)?
    \item[] Answer: \answerYes{}
    \item[] Justification: 
    Detailed experiment settings are provided in Section~\ref{sec:experiments} and Appendix~\ref{app:exp_details}.

\item {\bf Open access to data and code}
    \item[] Question: Does the paper provide open access to the data and code, with sufficient instructions to faithfully reproduce the main experimental results, as described in supplemental material?
    \item[] Answer: \answerNo{}
    \item[] Justification: The code is not publicly available at submission. We will make it open source upon acceptance.

\item {\bf Experimental setting/details}
    \item[] Question: Does the paper specify all the training and test details (e.g., data splits, hyperparameters, how they were chosen, type of optimizer) necessary to understand the results?
    \item[] Answer: \answerYes{}
    \item[] Justification: 
    Detailed experiment settings are provided in Section~\ref{sec:experiments} and Appendix~\ref{app:exp_details}.

\item {\bf Experiment statistical significance}
    \item[] Question: Does the paper report error bars suitably and correctly defined or other appropriate information about the statistical significance of the experiments?
    \item[] Answer: \answerNo{}
    \item[] Justification: 
    Due to computational constraints, we report results under fixed experimental settings without multiple runs or statistical significance tests.

\item {\bf Experiments compute resources}
    \item[] Question: For each experiment, does the paper provide sufficient information on the computer resources (type of compute workers, memory, time of execution) needed to reproduce the experiments?
    \item[] Answer: \answerYes{}
    \item[] Justification: 
    All experiments are conducted on a single NVIDIA H100 GPU.
    
\item {\bf Code of ethics}
    \item[] Question: Does the research conducted in the paper conform, in every respect, with the NeurIPS Code of Ethics \url{https://neurips.cc/public/EthicsGuidelines}?
    \item[] Answer: \answerYes{}
    \item[] Justification: 
    We adhere to the NeurIPS Code of Ethics.

\item {\bf Broader impacts}
    \item[] Question: Does the paper discuss both potential positive societal impacts and negative societal impacts of the work performed?
    \item[] Answer: \answerYes{}
    \item[] Justification: 
    The paper discusses how user-level traceability can help mitigate AI-generated content misuse and copyright infringement, while noting the risk of incorrect attribution.
    
\item {\bf Safeguards}
    \item[] Question: Does the paper describe safeguards that have been put in place for responsible release of data or models that have a high risk for misuse (e.g., pre-trained language models, image generators, or scraped datasets)?
    \item[] Answer: \answerNA{}
    \item[] Justification: 
    The paper does not release high-risk data or models that require safeguards.

\item {\bf Licenses for existing assets}
    \item[] Question: Are the creators or original owners of assets (e.g., code, data, models), used in the paper, properly credited and are the license and terms of use explicitly mentioned and properly respected?
    \item[] Answer: \answerYes{}
    \item[] Justification:
    We use MS-COCO, Stable-Diffusion-Prompts, SD3.5-medium, FLUX.1-dev, and MarkDiffusion under their respective licenses and terms of use.

\item {\bf New assets}
    \item[] Question: Are new assets introduced in the paper well documented and is the documentation provided alongside the assets?
    \item[] Answer: \answerNA{}
    \item[] Justification: 
    The paper does not release new assets at submission. Code and documentation will be provided upon acceptance.

\item {\bf Crowdsourcing and research with human subjects}
    \item[] Question: For crowdsourcing experiments and research with human subjects, does the paper include the full text of instructions given to participants and screenshots, if applicable, as well as details about compensation (if any)? 
    \item[] Answer: \answerNA{}.
    \item[] Justification: 
    The paper does not involve crowdsourcing or research with human subjects.

\item {\bf Institutional review board (IRB) approvals or equivalent for research with human subjects}
    \item[] Question: Does the paper describe potential risks incurred by study participants, whether such risks were disclosed to the subjects, and whether Institutional Review Board (IRB) approvals (or an equivalent approval/review based on the requirements of your country or institution) were obtained?
    \item[] Answer: \answerNA{}
    \item[] Justification: 
    The paper does not involve human subjects research and therefore does not require IRB approval.

\item {\bf Declaration of LLM usage}
    \item[] Question: Does the paper describe the usage of LLMs if it is an important, original, or non-standard component of the core methods in this research? Note that if the LLM is used only for writing, editing, or formatting purposes and does \emph{not} impact the core methodology, scientific rigor, or originality of the research, declaration is not required.
    \item[] Answer: \answerNA{}
    \item[] Justification: 
    LLMs were used only for language editing, such as grammar, spelling, and wording. The core method development does not involve LLMs as important, original, or non-standard components.

\end{enumerate}